\documentclass{article}

\PassOptionsToPackage{numbers, compress}{natbib}

\usepackage[main, preprint]{neurips_2026}

\usepackage[utf8]{inputenc} 
\usepackage[T1]{fontenc}    
\usepackage{url}            
\usepackage{booktabs}       
\usepackage{amsfonts}       
\usepackage{nicefrac}       
\usepackage{microtype}      

\usepackage{graphicx}
\usepackage{subcaption}
\usepackage{booktabs}
\usepackage{amsmath}
\usepackage{amssymb}
\usepackage{mathtools}
\usepackage{amsthm}
\usepackage{bm}
\usepackage{wrapfig}
\usepackage{fontawesome5}
\usepackage{algorithm}
\usepackage{algpseudocode}
\usepackage{pifont}
\usepackage{bbm}
\usepackage{multirow}
\usepackage[shortlabels]{enumitem}
\usepackage[dvipsnames]{xcolor}   
\usepackage[colorlinks=true, allcolors=NavyBlue]{hyperref}
\usepackage[capitalize,noabbrev]{cleveref}
\usepackage[textsize=tiny]{todonotes}
\usepackage[most]{tcolorbox}
\usepackage{listings}

\usepackage{amsmath,amsfonts,bm}

\def\eqref#1{equation~\ref{#1}}

\def\1{\bm{1}}

\def\va{{\bm{a}}}

\def\vh{{\bm{h}}}

\def\vo{{\bm{o}}}

\def\vu{{\bm{u}}}

\def\vz{{\bm{z}}}

\def\mI{{\bm{I}}}

\def\mZ{{\bm{Z}}}

\DeclareMathAlphabet{\mathsfit}{\encodingdefault}{\sfdefault}{m}{sl}
\SetMathAlphabet{\mathsfit}{bold}{\encodingdefault}{\sfdefault}{bx}{n}

\def\gC{{\mathcal{C}}}

\def\gL{{\mathcal{L}}}

\newcommand{\R}{\mathbb{R}}

\newcommand{\enc}{{\rm enc}_\theta}
\newcommand{\pred}{{\rm pred}_\phi}

\definecolor{predcolor}{HTML}{e74c3c}
\definecolor{anticolor}{HTML}{4a90d9}

\definecolor{codebg}{RGB}{248,248,248}
\definecolor{codecolor}{RGB}{140,142,144}

\definecolor{keyword}{RGB}{50, 106, 144}
\definecolor{string}{RGB}{215, 86, 68}
\definecolor{comment}{RGB}{130, 193, 139}
\definecolor{builtin}{RGB}{105, 201, 140}
\definecolor{var}{RGB}{105, 154, 187}
\definecolor{num}{RGB}{226, 136, 122}

\lstdefinestyle{pythonstyle}{
  language=Python,
  basicstyle= \fontfamily{fvm}\selectfont\small\color{codecolor},
  keywordstyle=\color{keyword}\bfseries,
  stringstyle=\color{string},
  commentstyle=\color{comment}\itshape,
  emph={obs,actions,lambd,next_emb,emb},
  emphstyle=\color{var},
  emph={[2]return, def, for, in},
  emphstyle={[3]\color{num}\bfseries},
  literate=
    {0}{{{\color{num}0}}}{1}
    {1}{{{\color{num}1}}}{1}
    {-1}{{{\color{num}-1}}}{2}
    {.0}{{{\color{num}.0}}}{2}
    {.1}{{{\color{num}.1}}}{2},
  showstringspaces=false,
  breaklines=true,
  frame=none,
  xleftmargin=1em,
}

\title{LeWorldModel: Stable End-to-End Joint-Embedding Predictive Architecture from Pixels}

\author{%
Lucas Maes\textnormal{*\textsuperscript{1}}~~ Quentin Le Lidec\textnormal{*\textsuperscript{2}}~~ Damien Scieur\textnormal{\textsuperscript{1,3}}~~ Yann LeCun\textnormal{\textsuperscript{2}}~~Randall Balestriero\textnormal{\textsuperscript{4}}
\\[8pt]
$^{1}$Mila \& Université de Montréal~~$^{2}$New York University~~$^{3}$Samsung SAIL ~~$^{4}$Brown University
}

\begin{document}

\renewcommand{\thefootnote}{}
\footnotetext{* Equal contribution. Correspondence to \texttt{lucas.maes@mila.quebec}}
\renewcommand{\thefootnote}{\arabic{footnote}}

\maketitle

\vspace{-2.5em}
\begin{center}
\href{https://le-wm.github.io}{%
  \tcbox[
    on line,
    colback=codebg,
    colframe=codebg,
    coltext=codecolor,
    boxrule=0.4pt,
    arc=4pt,
    boxsep=2pt,
    left=4pt, right=4pt, top=2pt, bottom=2pt
  ]{\fontfamily{fvm}\selectfont\small\faGlobe\enspace Website}}
\href{https://github.com/lucas-maes/le-wm}{%
  \tcbox[
    on line,
    colback=codebg,
    colframe=codebg,
    coltext=codecolor,
    boxrule=0.4pt,
    arc=4pt,
    boxsep=2pt,
    left=4pt, right=4pt, top=2pt, bottom=2pt
  ]{\fontfamily{fvm}\selectfont\small\faGithub\enspace Code}}
\end{center}
\vspace{2pt}

\begin{abstract}
Joint Embedding Predictive Architectures (JEPAs) offer a compelling framework for learning world models in compact latent spaces, yet existing methods remain fragile, relying on complex multi-term losses, exponential moving averages, pre-trained encoders, or auxiliary supervision to avoid representation collapse. In this work, we introduce LeWorldModel (LeWM), the first JEPA that trains stably end-to-end from raw pixels using only two loss terms: a next-embedding prediction loss and a regularizer enforcing Gaussian-distributed latent embeddings. This reduces tunable loss hyperparameters from six to one compared to the only existing end-to-end alternative. With ~15M parameters trainable on a single GPU in a few hours, LeWM plans up to $48\times$ faster than foundation-model-based world models while remaining competitive across diverse 2D and 3D control tasks. Beyond control, we show that LeWM's latent space encodes meaningful physical structure through probing of physical quantities. Surprise evaluation confirms that the model reliably detects physically implausible events. Code available \href{https://github.com/lucas-maes/le-wm}{here}.
\end{abstract}

\begin{figure}[h]
  \centering
  \includegraphics[width=\linewidth]{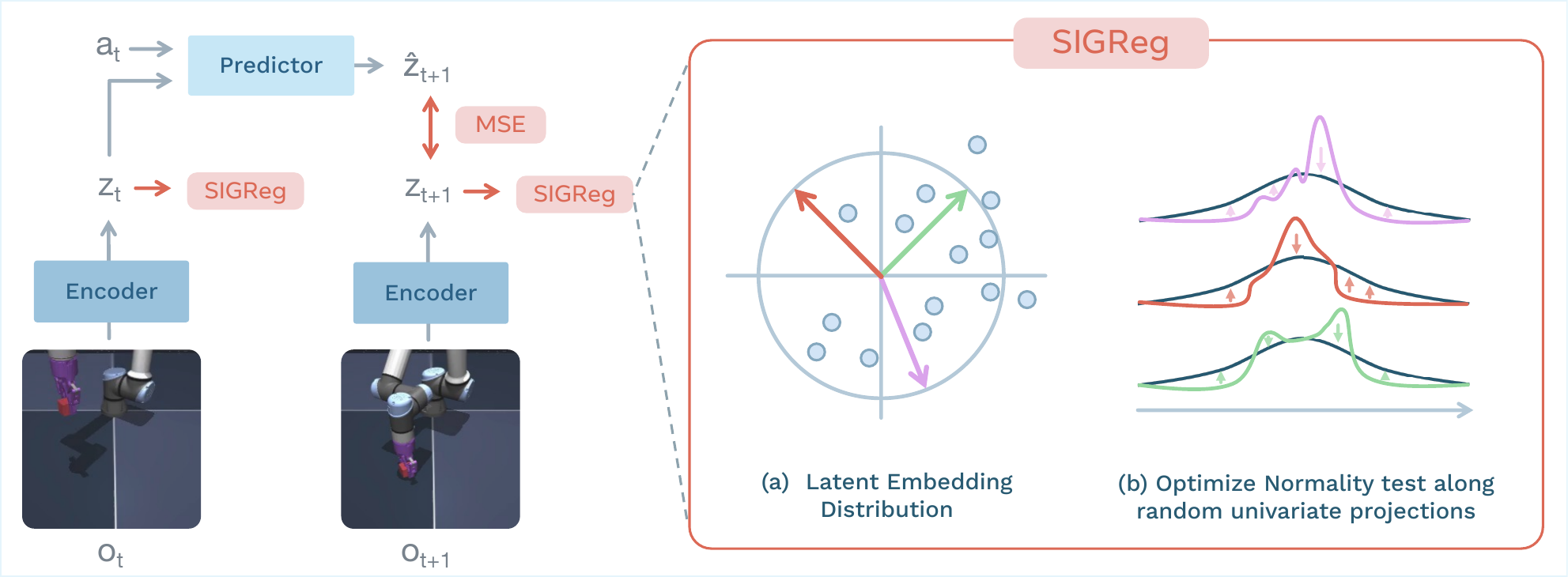}
  \caption{\small \textbf{LeWorldModel Training Pipeline.} Given frame observations $\vo_{1:T}$ and actions $\va_{1:T}$, the encoder maps frames into low-dimensional latent representations $\vz_{1:T}$. The predictor models the environment dynamics by autoregressively predicting the next latent state $\vz_{t+1}$ from the current latent state $\vz_t$ and action $\va_t$. The encoder and predictor are jointly optimized using a mean-squared error (MSE) prediction loss. LeWM does not rely on any training heuristics, such as stop-gradient, exponential moving averages, or pre-trained representations. To prevent trivial collapse, the SIGReg regularization term enforces Gaussian-distributed latent embeddings, promoting feature diversity. For tractability, latent embeddings are projected onto multiple random directions, and a normality test is applied to each one-dimensional projection. Aggregating these statistics encourages the full embedding distribution to match an isotropic Gaussian.}
  \label{fig:training-framework}
\end{figure}

\section{Introduction}

\begin{figure}
    \centering
    \includegraphics[width=0.9\linewidth]{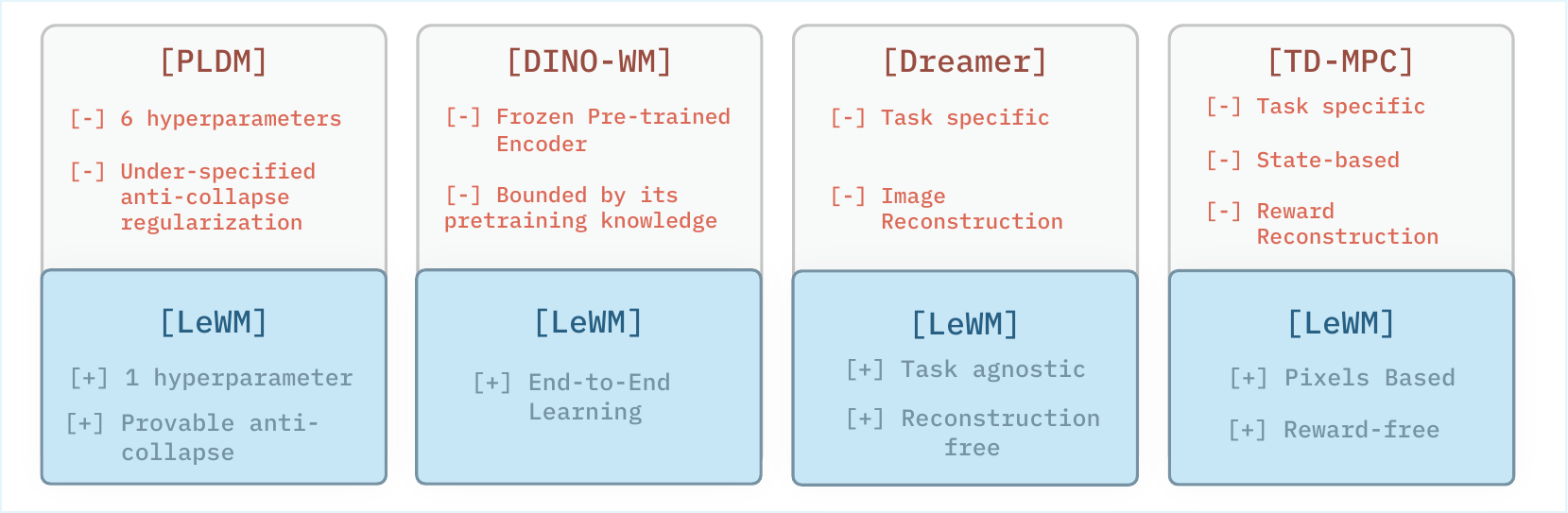}
    \caption{\small \textbf{Characteristics of latent world model approaches.} Methods are grouped by training paradigm. \textit{End-to-end} methods (PLDM) learn the encoder and predictor jointly from pixels without pre-trained representations or heuristics like stop-gradient or EMAs, but require many hyperparameters and lack collapse guarantees. \textit{Foundation-based} methods (DINO-WM) avoid collapse by freezing a pre-trained vision encoder, forgoing end-to-end learning. \textit{Task-specific} methods (Dreamer, TD-MPC) require reward signals or privileged state access. LeWM combines the strengths of all three: end-to-end, task-agnostic, pixel-based, reconstruction- and reward-free, with a single hyperparameter with provable anti-collapse guarantees.}
    \label{fig:wm}
    \vspace{-5mm}
\end{figure}

A central goal of artificial intelligence is to develop agents that acquire skills across diverse tasks and environments using a single, unified learning par adigm—one that operates directly from sensory inputs of its surroundings--without hand-engineered state representations or domain-specific calibration. Vision is ideally suited for this aim: cameras are inexpensive and scalable, and learning from pixels enables fully end-to-end training from raw sensory input to action \cite{levine2016end}.
World Models (WMs) are a powerful family of methods \cite{ha2018world} that learn to predict the consequences of actions in the environment. When successful, WMs allows agents to plan and to improve themselves solely form their model of the world, i.e., in imagination space. This is particularly valuable in the offline setting, where agents must learn from fixed datasets without environment interaction—leveraging the model to generate synthetic experience and evaluate counterfactual action sequences \cite{micheli2023transformers, hafner2025dreamerv4}. 

A recent popular approach for learning world models is the Joint Embedding Predictive Architecture (JEPA)~\cite{lecun2022path}. Instead of attempting to model every aspect of the environment, JEPA focuses on capturing the most relevant features needed to predict future states. Concretely, JEPA learns to encode observations into a compact, low-dimensional latent space and models temporal dynamics by predicting the latent representation of future observations.

However, despite their conceptual simplicity, existing JEPA methods are highly prone to collapse. In this failure mode, the model maps all inputs to nearly identical representations to trivially satisfy the temporal prediction objective leading to unusable representations. Preventing collapse is therefore one of the central challenges in training JEPA models. Many influential works have proposed methods to address this issue. Yet, these approaches typically rely on heuristic regularization, multi-objective loss functions, external sources of information, or architectural simplifications such as pre-trained encoders. In practice, these strategies often introduce additional instability or significantly increase training complexity (see App.~\ref{appendix:baselines}).

To overcome these limitations, we propose LeWorldModel (LeWM), the first method to learn a stable JEPA end-to-end from raw pixels without heuristics, principled, and simple (cf. Fig \ref{fig:training-framework}). We evaluate LeWM across a diverse set of manipulation, navigation, and locomotion tasks in both 2D and 3D environments. In addition, we probe its intuitive physical understanding through targeted probing and surprise-quantification evaluations in latent space. Overall, our key findings and contributions are:

\begin{itemize}
    \item We propose an end-to-end JEPA method for learning a latent world model from raw pixels on a single GPU.  The method relies on a simple and stable two-term objective that remains robust across architectures and hyperparameter choices, while enabling efficient logarithmic-time hyperparameter search.
    \item Our experiment demonstrates that LeWM achieves competitive control performance across diverse 2D and 3D tasks with only a compact 15M-parameter model, surpassing existing end-to-end JEPA-based approaches while remaining competitive with foundation-model-based world models at substantially lower cost, enabling planning up to $48\times$ faster.
    \item We evaluate physical understanding in the latent space through probing of physical quantities and a violation-of-expectation test for detecting unphysical trajectories.
\end{itemize}

\section{Related Work}

\begin{figure}[t]
    \centering
    \includegraphics[width=0.32\linewidth]{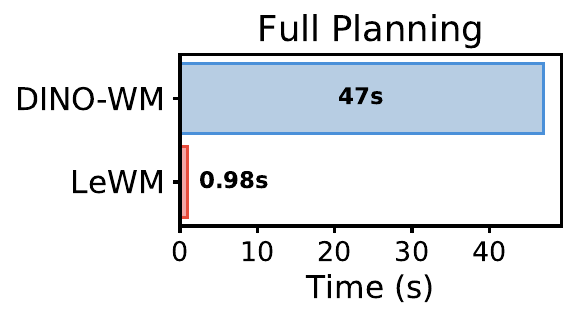}
    \includegraphics[width=0.32\linewidth]{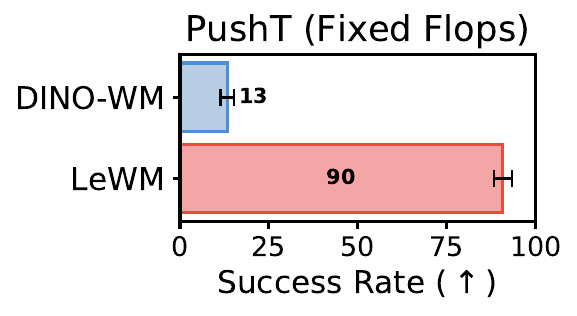}
    \includegraphics[width=0.32\linewidth]{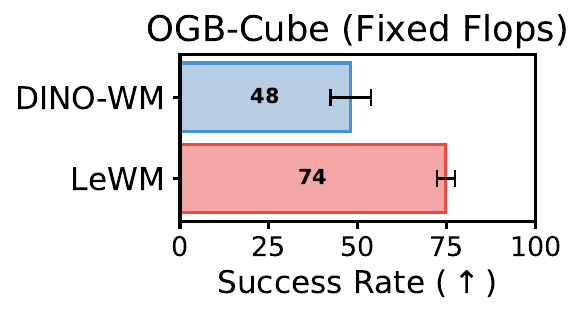}
    \caption{\small \textbf{Planning time and performance under fixed compute.}
    \textbf{Left:} Planning time comparison averaged over 50 runs. Encoding observations with $\sim200\times$ fewer tokens than DINO-WM allows LeWM to achieve planning speeds comparable to PLDM while being up to $\sim50\times$ faster than DINO-WM.
    \textbf{Center–Right:} Planning performance under the same computational budget (fixed FLOPs). LeWM significantly outperforms DINO-WM on Push-T (center) and OGBench-Cube (right). See App.~\ref{appendix:details} for planning setup details.
    }
    \label{fig:planning-time}
    \vspace{-6mm}
\end{figure}

\noindent \textbf{World Models} aim to learn predictive models of environment dynamics from data, enabling agents to reason about future states in imagination. A prominent class of WMs consists of \emph{generative} approaches that explicitly model environment dynamics in pixel space. These action-conditioned generative models act as learned simulators by producing future observations conditioned on past states and actions.
Generative world models have been successfully applied to simulate existing game-like environments. For example, IRIS~\cite{micheli2023transformers}, DIAMOND~\cite{alonso2024diffusion}, $\Delta$-IRIS~\cite{micheli2024efficient}, OASIS~\cite{oasis2024}, and DreamerV4~\cite{hafner2025dreamerv4} model environments such as Minecraft, Counter-Strike, and Crafter, improving policy sample efficiency in reinforcement learning. Other methods generate entirely new interactive simulators, e.g., Genie~\cite{bruce2024geniegenerativeinteractiveenvironments} and HunyuanWorld~\cite{hunyuanworld2025tencent}, while learned simulators have also been applied to robot policy evaluation~\cite{quevedo2025worldgym}.
Importantly, many generative WMs assume access to datasets containing reward signals, enabling joint modeling of dynamics and value-relevant information for downstream reinforcement learning. In contrast, we focus on the reward-free setting, corresponding to the setup considered in the JEPA line of work, which aims at learning generic, task-agnostic world models from observational data without relying on reward supervision.

\noindent \textbf{JEPA} is a framework for learning world models that predict the dynamic evolution of a system in a compact, low-dimensional latent space. Since their introduction by \citet{lecun2022path}, JEPA methods have evolved considerably, differing mainly in their target tasks and in the strategies used to learn non-collapsing representations. One prominent line of work applies JEPA to self-supervised representation learning by predicting the latent embeddings of masked input patches. Examples include I-JEPA~\cite{assran2023self} for images, V-JEPA~\cite{bardes2023v, assran2025v} for videos, and Echo-JEPA and Brain-JEPA~\cite{dong2024brainjepa, munim2026echojepalatentpredictivefoundation} for medical data. These approaches typically employ an exponential moving average (EMA) of the target encoder together with stop-gradient (SG) updates to stabilize training and prevent representation collapse. However, the theoretical understanding of EMA and SG remains limited, as they do not in general correspond to the minimization of a well-defined objective~\cite{ponce2026dual}. A second line of work uses the JEPA recipe for action-conditioned latent world modeling. Some approaches rely on pretrained encoders to obtain representations~\cite{assran2025v, zhou2025dino-wm, goswami2025osviwmoneshotvisualimitation,nam2026causaljepalearningworldmodels}. This avoids collapse but limits the expressivity of representation to the pretrained encoder used. In contrast, PLDM~\cite{sobal2022jointembeddingpredictivearchitectures,sobal2025stresstesting} learns representations end-to-end using VICReg~\cite{bardes2022vicreg} with additional regularization terms, at the cost of known training instabilities and scalability limitations~\citep{balestriero2022contrastive}. Several works further improve stability by incorporating auxiliary signals or architectural components, such as proprioceptive inputs or action decoders~\cite{zhou2025dino-wm, goswami2025osviwmoneshotvisualimitation}. In this work, we propose a stable method for training end-to-end JEPAs directly from raw pixels using a simple two-term loss: a predictive objective on future embeddings and a regularization objective that enforces Gaussian-distributed embeddings~\citep{balestriero2025lejepa}.

\noindent \textbf{Planning with Latent Dynamics.}
World Models~\cite{ha2018recurrent} pioneered learning policies directly from compact latent representations of high-dimensional observations. Some works leverage learned latent dynamics models to train policies using reinforcement learning~\cite{Hafner2020Dream,hafner2020dreamerv2,hafner2023dreamerv3,hafner2025dreamerv4}. In these approaches, the generative world model acts as a simulator in which trajectories are rolled out in imagination, allowing policy optimization to occur largely in imagination in latent space. Once training is complete, the policy is executed directly, and the world model is no longer required at test time. More recent works instead perform planning directly in the latent space at test time using Model Predictive Control (MPC)~\cite{testud1978model,Hansen2022tdmpc,hansen2024tdmpc,bar2025navigationworldmodels,zhou2025dino-wm,sobal2025stresstesting}.

\begin{figure}[t]
    \centering
    \includegraphics[width=\linewidth]{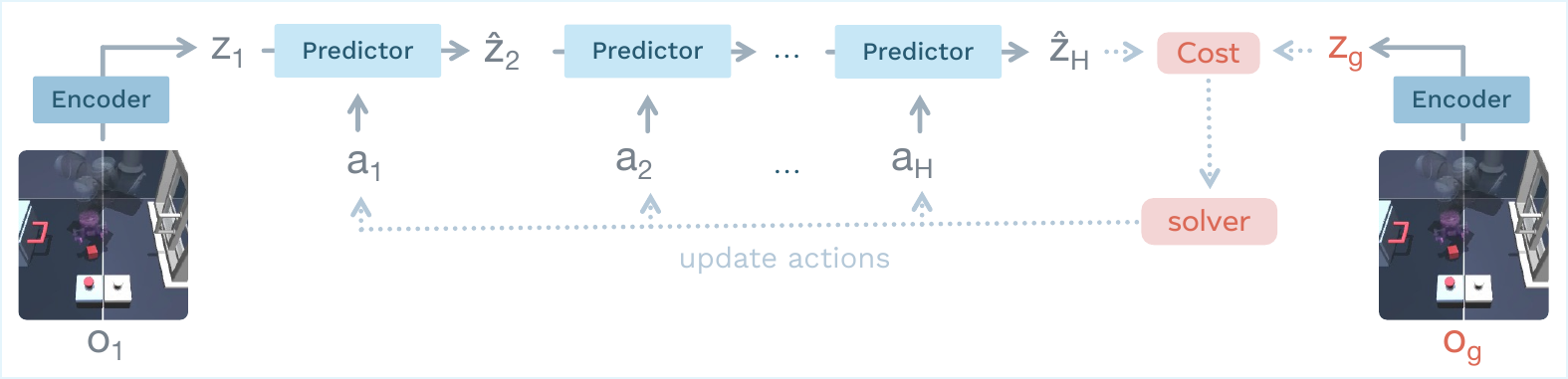}
    \caption{\small \textbf{Latent Planning with LeWorldModel .} Given an initial observation $\vo_1$ and a goal $\vo_g$, the world model learned in Fig.~2 performs planning in the LeWM latent space. The initial state embedding $\vz_1$ and the goal embedding $\vz_g$ are obtained from the encoder. The predictor then rolls out future latent states up to a horizon $H$. A latent cost between the final predicted state and the goal embedding guides a solver to optimize the action sequence. This prediction–optimization loop is repeated until convergence to a good plan candidate.}
    \label{fig:lewm}
    \vspace{-6mm}
\end{figure}

\section{Method: LeWorldModel}

In this section, we introduce LeWorldModel (LeWM). We first describe the streamlined training procedure used to learn the latent world model from offline data, including the dataset, model architecture, and training objective. We then explain how the learned model can be leveraged for decision making through latent planning using model predictive control (MPC).

\subsection{Learning the Latent World Model}

\paragraph{Offline Dataset.}
We consider a fully offline and reward-free setting. LeWorldModel is trained solely from unannotated trajectories of observations and actions, without access to reward signals or task specifications. This setup aligns with the JEPA line of work \cite{zhou2025dino-wm,assran2025v}, which aims to learn generic, task-agnostic world models from observational data. Our objective is not to optimize behavior for a specific task, but to learn representations that capture environment dynamics and can later be controlled or adapted to a diverse set of tasks.

The training data consists of trajectories of length $T$ composed of raw pixel observations $\vo_{1:T}$ and associated actions $\va_{1:T}$. Trajectories are collected offline from behavior policies with no optimality requirements; they may be pseudo-expert or exploratory, as long as they sufficiently cover the environment dynamics.
Additional implementation details (batch size, resolution, and sub-trajectory construction) are provided in App.~\ref{appendix:details}.

\paragraph{Model Architecture.} LeWM is built upon two components: an encoder and a predictor. The encoder maps a given frame observation $\vo_t$ into a compact, low-dimensional latent representation $\vz_t$. 
The predictor models the environment dynamics in latent space by predicting the embedding of the next frame observation $\hat{\vz}_{t+1}$ given the latent embedding $\vz_t$ and an action $\va_t$.

\begin{equation}
\begin{aligned}
\text{Encoder:} \quad & \vz_t = \enc(\vo_t) \\
\text{Predictor:} \quad & \hat{\vz}_{t+1} = \pred(\vz_t, \va_t)
\end{aligned}
\tag{LeWM}
\end{equation}

The encoder is implemented as a Vision Transformer (ViT)~\cite{dosovitskiy2020image}. Unless otherwise specified, we use the tiny configuration ($\sim$5M parameters) with a patch size of 14, 12 layers, 3 attention heads, and hidden dimensions of 192. The observation embedding $\vz_t$ is constructed from the \texttt{[CLS]} token embedding of the last layer, followed by a projection step. The projection step maps the \texttt{[CLS]} token embedding into a new representation space using a 1-layer MLP with Batch Normalization~\cite{ioffe2015batchnormalizationacceleratingdeep}. This step is necessary because the final ViT layer applies a Layer Normalization~\cite{ba2016layer}, which prevents our anti-collapse objective from being optimized effectively. 

The predictor is a transformer with 6 layers, 16 attention heads, and 10\% dropout ($\sim$10M parameters). Actions are incorporated into the predictor through Adaptive Layer Normalization (AdaLN)~\cite{peebles2023scalable} applied at each layer. The AdaLN parameters are initialized to zero to stabilize training and ensure that action conditioning impacts the predictor training progressively. The predictor takes as input a history of $N$ frame representations and predicts the next frame representation auto-regressively with temporal causal masking to avoid looking at future embeddings. The predictor is also followed by a projector network with the same implementation as the one used for the encoder.
All components of our world model are learned jointly using the loss described in the following paragraph.

\paragraph{Training Objective.}
Our objective is to learn latent representations useful for predicting the future, i.e., modeling the environment dynamics. {\rm LeWorldModel} training objective is the sum of two terms: a prediction loss and a regularization loss. The prediction loss $\mathcal{L}_{\rm pred}$ (teacher-forcing) computes the error between the predicted embedding of consecutive time-steps:

\begin{equation}
    \mathcal{L}_{\rm pred} \triangleq  \|\hat{\vz}_{t+1} - {\vz}_{t+1}\|^2_2, 
    \quad\quad 
    \hat{\vz}_{t+1} = {\rm pred}_\phi(\vz_t, \va_t).
    \label{eq:l2}
\end{equation} 
Through the prediction loss, the encoder is incentivized to learn a predictable representation for the predictor.

However, if alone, the loss in Eq.~\ref{eq:l2} leads to representation collapse, yielding a trivial solution in which the encoder maps all inputs to a constant representation. To prevent this behavior, we introduce an anti-collapse regularization term that promotes feature diversity in the embedding space. Specifically, we adopt the Sketched-Isotropic-Gaussian Regularizer (SIGReg)~\cite{balestriero2025lejepa} due to its simplicity, scalability, and stability. SIGReg encourages the latent embeddings to match an isotropic Gaussian target distribution.

Let $\mZ \in \R^{N \times B \times d}$ denote the tensor of latent embeddings collected over the history length $N$, the batch size $B$, and where $d$ denotes the embedding dimension. Assessing normality directly in high-dimensional spaces is challenging, as most classical normality tests are designed for univariate data and do not scale reliably with dimensionality. SIGReg circumvents this limitation by projecting embeddings onto $M$ random unit-norm directions $\vu^{(m)} \in \mathbb{S}^{d-1}$ and optimizing the univariate Epps--Pulley~\cite{epps1983test} test statistic $T(\cdot)$ along the resulting one-dimensional projections $\vh^{(m)} = \mZ \vu^{(m)}$, as illustrated in Fig.\ref{fig:training-framework}. By the Cramér--Wold theorem~\cite{cramer1936some}, matching all one-dimensional marginals is equivalent to matching the full joint distribution.

\begin{equation}
    {\rm SIGReg}(\mZ) \triangleq  \frac{1}{M} \sum_{m=1}^M T(\vh^{(m)}).
\end{equation}

Additional details on SIGReg and the definition of the Epps--Pulley statistical test are provided in appendix \ref{appendix:sigreg}.

The complete LeWM training objective is defined as:
\begin{equation}
\gL_{\rm LeWM} \triangleq \gL_{\rm pred} + \lambda\,{\rm SIGReg}(\mZ).
\end{equation}

The method introduces only two training hyperparameters: the number of random projections $M$ used in SIGReg and the regularization weight $\lambda$. Unless otherwise specified, we use $M = 1024$ projections and $\lambda = 0.1$. In practice, we observe that the number of projections has negligible impact on downstream performance (see Sec.~\ref{sec:exp_control} and App.~\ref{appendix:ablations}), making $\lambda$ the only effective hyperparameter to tune. This greatly simplifies hyperparameter selection, as $\lambda$ can be efficiently optimized using a simple bisection search with logarithmic complexity. We do not employ stop-gradient, exponential moving averages, or additional stabilization heuristics. Gradients are propagated through all components of the loss, and all parameters are optimized jointly in an end-to-end manner, resulting in a streamlined and easy-to-implement training procedure. The training logic is summarized in Alg.~\ref{alg:train-alg}.

\subsection{Latent Planning}

At inference time, we perform trajectory optimization in our world model latent space, as illustrated in Fig.\ref{fig:lewm}. Given an initial observation $\vo_1$, we initialize a candidate action sequence randomly and iteratively rollout predicted latent states up to a planning horizon $H$. The model predicts latent transitions according to
\begin{equation*}
\hat{\vz}_{t+1} = \pred(\hat{\vz}_t, \va_t), \quad
\hat{\vz}_1 = \enc(\vo_1),
\end{equation*}
Planning is performed by optimizing the action sequence to minimize a terminal latent goal-matching objective,
\begin{equation}
\gC(\hat{\vz}_H) = \| \hat{\vz}_H - \vz_g \|_2^2, \quad
\vz_g = \enc(\vo_g),
\end{equation}
where $\hat{\vz}_H$ is the predicted latent state at the end of the rollout and $\vz_g$ is the latent embedding of the goal observation $\vo_g$. The world model parameters remain fixed during planning. This procedure corresponds to a finite-horizon optimal control problem,
\begin{equation}
\va^*_{1:H} = \arg\min_{\va_{1:H}} \gC(\hat{\vz}_H),
\label{eq:argmin_plan}
\end{equation}
which we solve using the Cross-Entropy Method (CEM)~\cite{rubinstein2004cross}, a sampling method that iteratively selects the best plan and updates the parameters of the sampling distribution with the statistics of the best plans. The planning horizon $H$ trades off long-term lookahead against increased computational cost and model bias. In particular, auto-regressive rollouts accumulate prediction errors as the horizon grows, which can deteriorate the quality of the optimized action sequence. To mitigate this effect, we adopt a Model Predictive Control (MPC) strategy: only the first $K$ planned actions are executed before replanning from the updated observation.
We provide more details on the planning strategy in appendix~\ref{appendix:details}.

\section{Latent Planning Performance} \label{sec:exp_control}
\subsection{Planning evaluation setup}

\begin{figure}[t]
    \centering
    \includegraphics[width=0.9\linewidth]{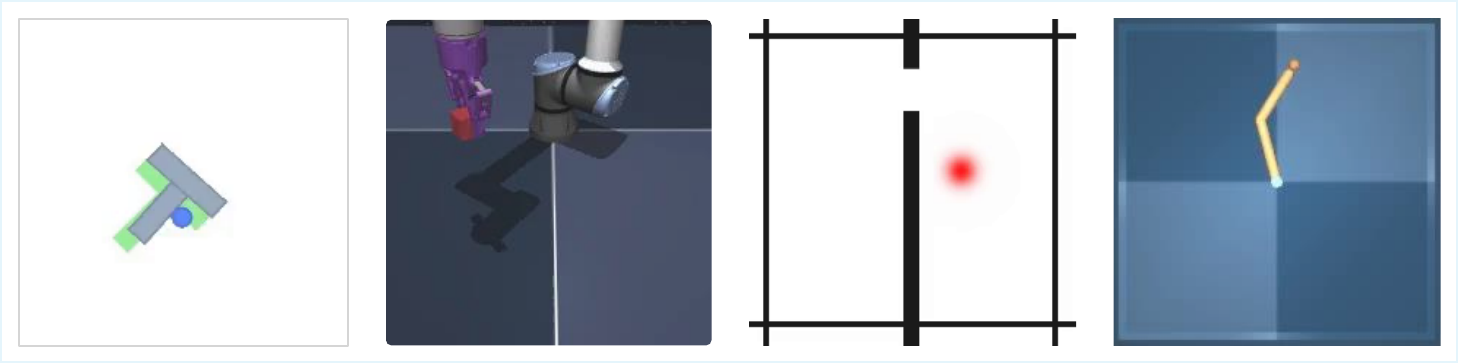}
    \caption{\small Environments used for evaluation. \textbf{Left:} Push-T, a 2D manipulation task where the agent must push a block toward a target configuration, commonly used as a robotics benchmark. \textbf{Center (1):} OGBench-Cube, a visually richer 3D manipulation environment where a robotic arm interacts with a cube to reach a target position. \textbf{Center (2):} Two-Room, a simple 2D navigation environment where an agent moves between rooms to reach target positions. \textbf{Right:} Reacher, a task where a 2-joint arm needs to reach a target configuration in a 2D plane. All environments have a continuous action space. More details on environment and datasets are available in appendix~\ref{appendix:envs}.}
    \label{fig:envs}
    \vspace{-6mm}
\end{figure}

\paragraph{Environments.} We evaluate LeWM on a diverse set of tasks, including navigation, motion planning and manipulation, in both two- and three-dimensional environments, all illustrated in Fig.~\ref{fig:envs}. We provide more details on dataset generation and environments in App.~\ref{appendix:envs}.

\paragraph{Baselines.} We compare the performance of LeWM against several baselines: DINO-WM~\citep{zhou2025dino-wm} and PLDM~\citep{sobal2025stresstesting}, two state-of-the-art JEPA-based methods; a goal-conditioned behavioral cloning policy (GCBC); and two goal-conditioned offline reinforcement learning algorithms, GCIVL and GCIQL~\citep{kostrikov2021offline}.
Among these baselines, PLDM is the closest to our setup, as it also learns a world model end-to-end directly from pixel observations. However, it relies on a seven-term training objective derived from the VICReg~\citep{bardes2022vicreg} criterion, which introduces training instability and increases the complexity of hyperparameter tuning. DINO-WM, in contrast, models dynamics using DINOv2~\cite{oquab2024dinov} as feature encoder to mitigate representation collapse, but its original formulation additionally incorporates other modalities, such as proprioceptive inputs; for a fair comparison, unless specified otherwise, we exclude proprioceptive information from DINO-WM. Additional implementation details for the baselines (App.~\ref{appendix:baselines}) and evaluation settings (App.~\ref{appendix:control_details}) are provided in the appendix. For each method, we keep the hyperparameters fixed across all environments.

\subsection{Towards Efficient Planning with WMs}

\begin{figure}[t]
    \centering
    
    \begin{subfigure}{0.24\linewidth}
        \centering
        \includegraphics[width=\linewidth]{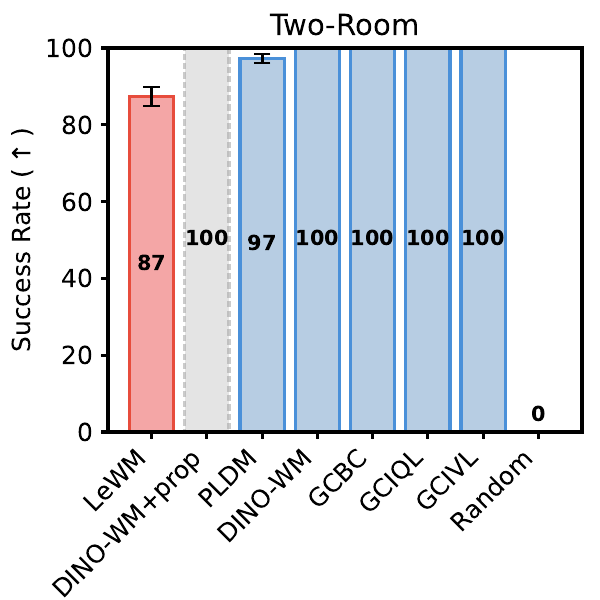}
    \end{subfigure}
    \hfill
    \begin{subfigure}{0.24\linewidth}
        \centering
        \includegraphics[width=\linewidth]{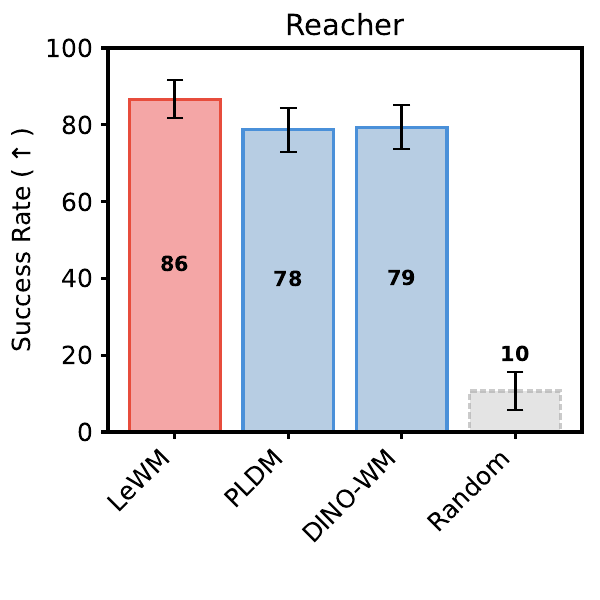}
    \end{subfigure}
    \hfill
    \begin{subfigure}{0.24\linewidth}
        \centering
        \includegraphics[width=\linewidth]{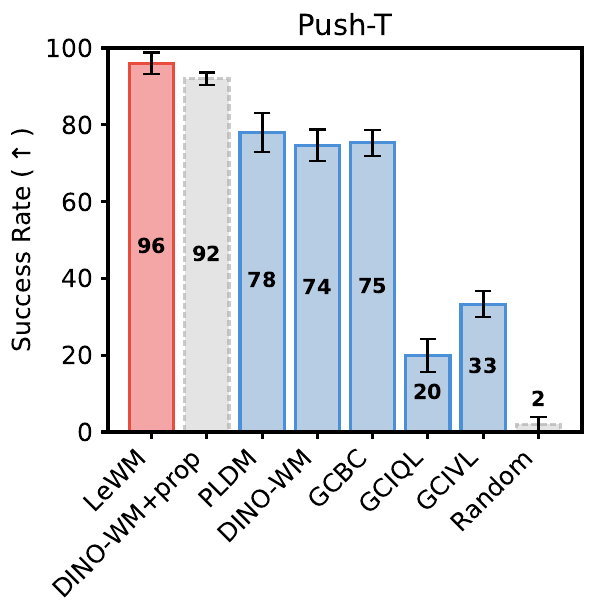}
    \end{subfigure}
    \hfill
    \begin{subfigure}{0.24\linewidth}
        \centering
        \includegraphics[width=\linewidth]{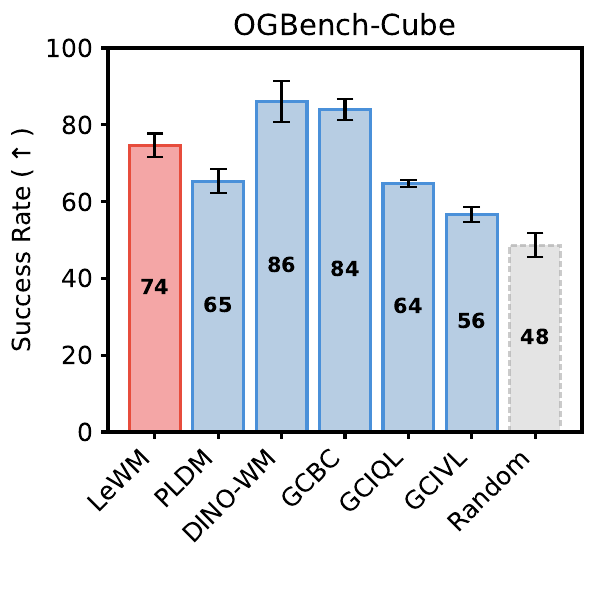}
    \end{subfigure}

    \caption{\small \textbf{Planning performance across environments.} 
    Results are shown for \textit{Two-Room} (left), \textit{Reacher} (center 1), \textit{PushT} (center-2) and \textit{OGBench-Cube} (right). 
    LeWM consistently outperforms PLDM and DINO-WM on Push-T and Reacher. On OGBench-Cube, DINO-WM slightly outperforms LeWM, possibly due to the higher visual complexity and the 3D nature of the environment, which makes encoder training more challenging. In the simpler Two-Room environment, PLDM and DINO-WM outperform LeWM, which may be explained by the SIGReg regularization encouraging a Gaussian distribution in a high-dimensional latent space, while the intrinsic dimensionality of the environment is much lower.}
    \label{fig:ctrl-all}
    \vspace{-6mm}
\end{figure}

We report planning performance in Fig.~\ref{fig:ctrl-all}. LeWM outperforms PLDM on the more challenging planning tasks, achieving an 18\% higher success rate on PushT, while remaining competitive with DINO-WM. Notably, on PushT, LeWM (pixels-only) surpasses DINO-WM even when DINO-WM has access to additional proprioceptive information, demonstrating LeWM’s ability to capture underlying task-relevant quantities. Interestingly, LeWM performs worse on the simplest environment, Two-Room. A possible explanation is that the low diversity and low intrinsic dimensionality of this dataset make it difficult for the encoder to match the isotropic Gaussian prior enforced by SIGReg in a high-dimensional latent space, which may lead to a less structured latent representation. This highlights a potential limitation of the SIGReg regularization in very low-complexity environments.

Moreover, when comparing planning speedups (Fig.~\ref{fig:planning-time}), LeWM achieves a $48\times$ faster planning time, with the full planning completing in under one second while preserving competitive performance across tasks. This planning time remains consistent across environments for a fixed planning setup, narrowing the gap toward real-time control.

\subsection{Towards Stable Training of World Models}

\paragraph{Ablations.} We perform ablations on several design choices of LeWM. First, we analyze the sensitivity of SIGReg to its internal parameters, namely the number of random projections and the number of integration knots. The performance is largely unaffected by these quantities, indicating that they do not require careful tuning. As a result, the regularization weight $\lambda$ remains the only effective hyperparameter. Since only a single hyperparameter needs to be tuned, grid search can be performed efficiently using a simple bisection strategy ($\mathcal{O}(\log n)$), whereas PLDM requires search in polynomial time ($\mathcal{O}(n^6)$). We also study the effect of the embedding dimensionality. While the representation dimension must be sufficiently large for the method to perform well, performance quickly saturates beyond a certain threshold, suggesting that the approach is robust to the precise choice of encoder capacity. Additionally, we examine the impact of the encoder architecture by replacing the default ViT encoder with a ResNet-18 backbone (Tab.~\ref{tab:encoder-arch}). LeWM achieves competitive performance with both architectures, indicating that it is largely agnostic to the choice of vision encoder. Details on all ablations are available in App.~\ref{appendix:ablations}.

\paragraph{Training Curves.}We report the training loss curves on PushT for LeWM in Fig.~\ref{fig:lewm-loss} and PLDM in Fig.~\ref{fig:pldm-loss}. The two-term objective of LeWM exhibits smooth and monotonic convergence: the prediction loss decreases steadily while the SIGReg regularization term drops sharply in the early phase of training before plateauing, indicating that the latent distribution quickly approaches the isotropic Gaussian target. In contrast, PLDM's seven-term objective displays noisy and non-monotonic behavior across several of its loss components. These observations highlight a key advantage of LeWM: by reducing the training objective to only two well-behaved terms, the training becomes significantly more stable, removing the need to balance competing gradients from multiple regularizers.

\section{Quantifying Physical Understanding in LeWM}

In this section, we evaluate the quality of the dynamics captured by LeWM’s latent space, either by learning to extract physical quantities from latent embeddings or by measuring the world model’s ability to detect changes in physics.

\subsection{Physical Structure of the Latent Space}

\paragraph{Probing physical quantities.} As a first measure of physical understanding, we evaluate which physical quantities are recoverable from LeWM’s latent representations. We train both linear and non-linear probes to predict physical quantities of interest from a given embedding. Results on the Push-T environment are reported in Tab.~\ref{tab:probe-pusht}. Our method consistently outperforms PLDM while remaining competitive with representations produced by large pretrained models such as DINOv2. We provide probing results on other environments in App.~\ref{appendix:probing}.

\begin{table}[h]
\centering
\caption{\small \textbf{Physical latent probing results on Push-T.} LeWM consistently
outperforms PLDM while remaining competitive with DINO-WM. The strong probing
performance of DINO-WM on certain properties may stem from its foundation-model
pretraining: the DINOv2 encoder is trained on two orders of magnitude more data
($\sim$124M images) spanning a far more diverse
distribution, which likely allows it to capture some physical properties in its
embeddings by default.}
{\footnotesize
\begin{tabular}{l l c c c c}
\toprule
& & \multicolumn{2}{c}{\textbf{Linear}} & \multicolumn{2}{c}{\textbf{MLP}}\\
\cmidrule(lr){3-4} \cmidrule(lr){5-6}
\textbf{Property} & \textbf{Model} & MSE $\downarrow$ & r $\uparrow$ & MSE $\downarrow$ & r $\uparrow$\\
\toprule
\multirow{3}{*}{Agent Location}
  & DINO-WM & $1.888 \pm 0.500$ & $\mathbf{0.977}$ & $\mathbf{0.003 \pm 0.022}$ & $\mathbf{0.999}$\\
  & PLDM    & $0.090 \pm 0.311$ & $0.955$ & $0.014 \pm 0.119$ & $0.993$\\
  & LeWM    & $\mathbf{0.052 \pm 0.149}$ & $0.974$ & $0.004 \pm 0.056$ & $0.998$\\
\midrule
\multirow{3}{*}{Block Location}
  & DINO-WM & $\mathbf{0.006 \pm 0.007}$ & $\mathbf{0.997}$ & $0.002 \pm 0.003$ & $\mathbf{0.999}$\\
  & PLDM    & $0.122 \pm 0.341$ & $0.938$ & $0.011 \pm 0.066$ & $0.994$\\
  & LeWM    & $0.029 \pm 0.073$ & $0.986$ & $\mathbf{0.001 \pm 0.006}$ & $\mathbf{0.999}$\\
\midrule
\multirow{3}{*}{Block Angle}
  & DINO-WM & $\mathbf{0.050 \pm 0.101}$ & $\mathbf{0.979}$ & $\mathbf{0.009 \pm 0.052}$ & $\mathbf{0.995}$\\
  & PLDM    & $0.446 \pm 0.625$ & $0.745$ & $0.056 \pm 0.184$ & $0.972$\\
  & LeWM    & $0.187 \pm 0.359$ & $0.902$ & $0.021 \pm 0.139$ & $0.990$\\
\bottomrule
\end{tabular}%
}
\label{tab:probe-pusht}
\vspace{-3mm}
\end{table}
\begin{figure}[t]
    \centering
    \includegraphics[width=0.9\linewidth]{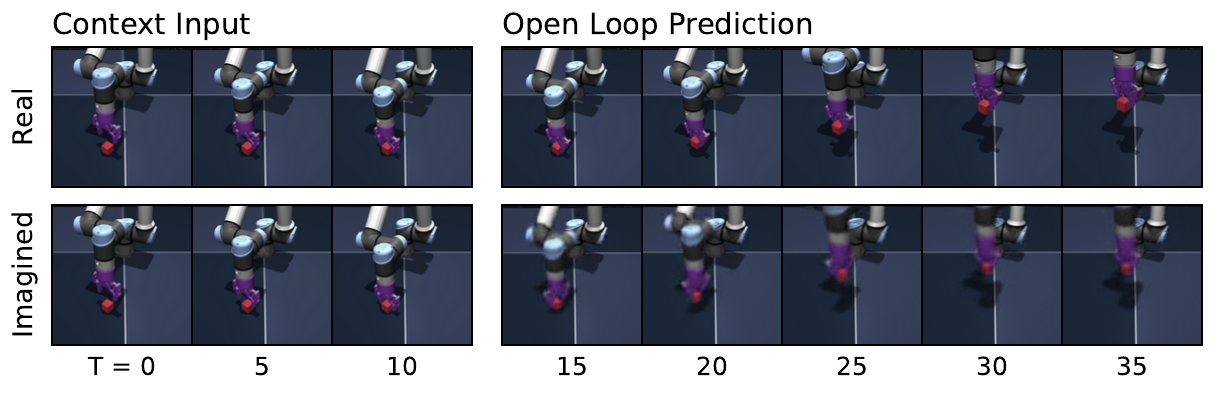}
    \caption{\small\textbf{Predictor rollout on OGBench-Cube}. We visualize decoded latent plans from LeWM given context and action sequences. Each rollout encodes three image observations as context, then autoregressively generates future latents conditioned on the actions in an open-loop fashion. Latents are decoded via a decoder trained \textit{a posteriori}. The imagined rollout suggests that latent representations capture the global scene structure while finer details like end-effector angle are not fully preserved. Additional rollouts for Push-T and OGBench-Cube are provided in Fig.~\ref{fig:rollout-2}.}
    \label{fig:rollout}
    \vspace{-5mm}
\end{figure}

\paragraph{Decoding Latent Space.}
To further assess the information captured in the latent representation, we report in Fig.~\ref{fig:decoder-vis} images produced by a decoder trained to reconstruct pixel observations from a single latent embedding (192 dim) during training. Although reconstruction is never used during training, the decoder is able to recover the visual scene from the learned representation, confirming that the low-dimensional and compact latent space retains sufficient information about the underlying physical state. Details on the decoder architecture are provided in App.~\ref{appendix:details}.

\paragraph{Visualizing Latent Space.} We further visualize the structure of the latent space using t-SNE. Fig.~\ref{fig:tsne_pusht_lejepa} provides a qualitative visualization of the latent space in the \textit{PushT} environment. The visualization suggests that the learned representation captures the spatial structure of the environment, preserving neighborhood relationships and relative positions in the latent space.

\paragraph{Temporal Latent Path Straightening.} Inspired by the temporal straightening hypothesis from neuroscience \citep{henaff2019perceptual} and recent work from ~\cite{interno2025aigenerated,wang2026temporal}, we measure the cosine similarity between consecutive latent velocity vectors throughout training (Eq.~\ref{eq:path-straightening}). We find that LeWM's latent trajectories become increasingly straight on PushT over training as a purely emergent phenomenon, without any explicit regularization encouraging this behavior, cf. Fig.~\ref{fig:tmp-straight}. Remarkably, LeWM achieves higher temporal straightness than PLDM, despite PLDM employing a dedicated temporal smoothness regularization term. We detail our findings in App.~\ref{appendix:tmp-straight}.

\subsection{Violation-of-expectation Framework} \label{subsec:voe}

\begin{figure}
    \centering
    \includegraphics[width=0.32\linewidth]{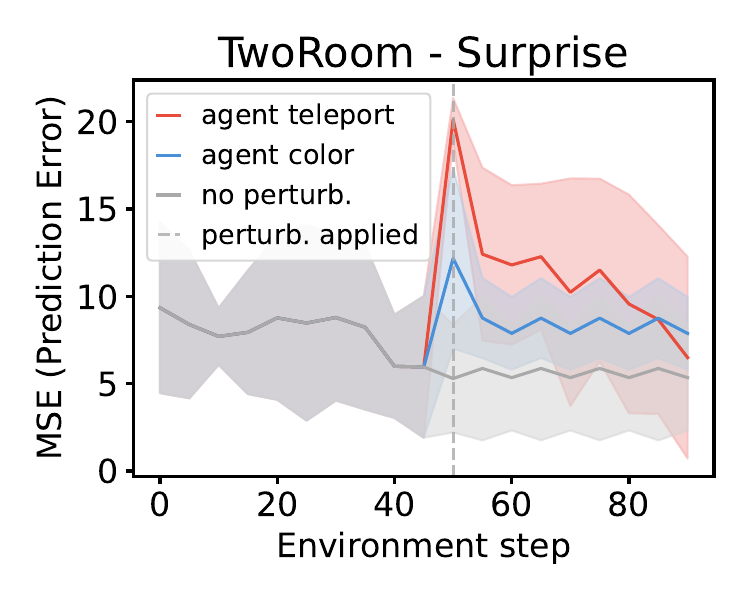}
    \includegraphics[width=0.32\linewidth]{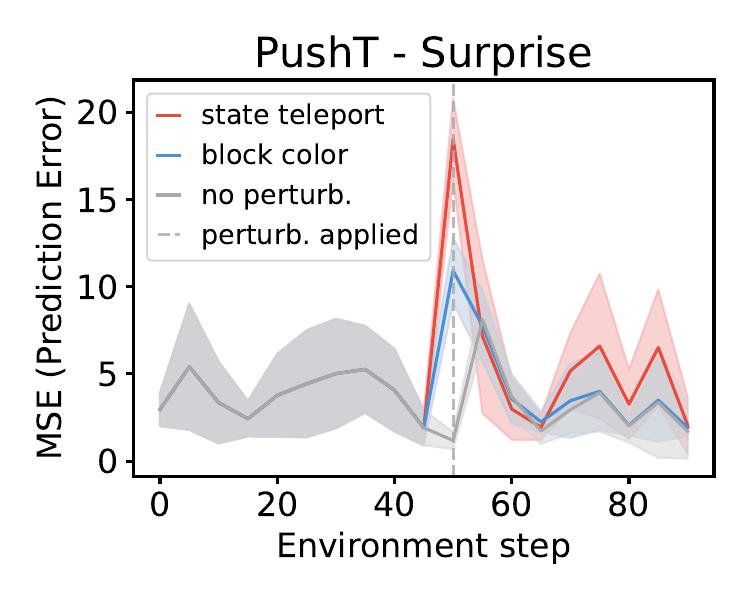}
    \includegraphics[width=0.32\linewidth]{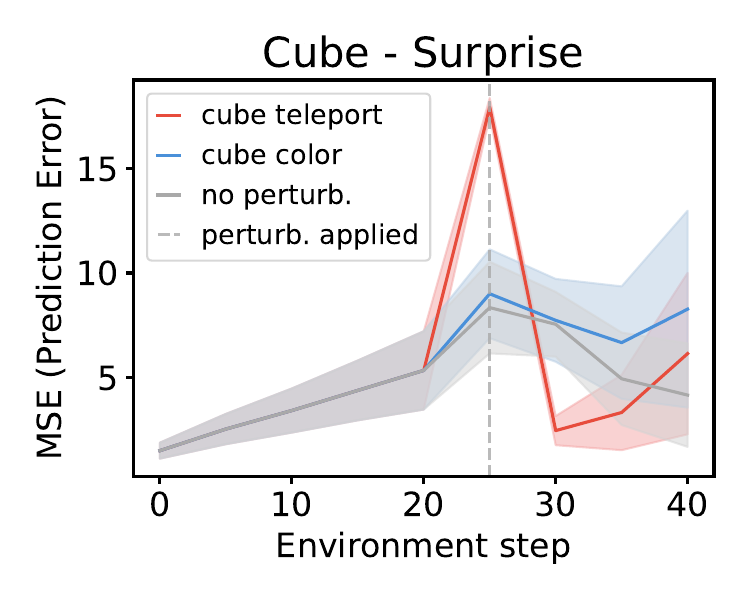}

    \caption{\small\textbf{Violation-of-expectation evaluation across three environments.} Each plot shows LeWM surprise along three trajectories: an unperturbed reference, a visually perturbed trajectory with abrupt object color change, and a physically perturbed trajectory where objects teleport to random positions. Teleportation violates physical continuity and produces a pronounced surprise spike, while the unperturbed trajectory stays at a low baseline. The increase is significant for teleportation across all environments (paired t-test, $p<0.01$) but weaker and non-significant for color changes, indicating greater sensitivity to physical than visual perturbations. Environments, left to right: \textit{TwoRoom}, \textit{PushT}, \textit{OGBench Cube}.}
    
    \label{fig:voe_surprise_lejepa}
    \vspace{-6mm}
\end{figure}

Another approach to quantifying physical understanding is the ability to detect violations of the learned world model. Inspired by the violation-of-expectation (VoE) paradigm used in developmental psychology and recently adopted in machine learning \cite{margoni2024violation, garrido2025intuitive, bordes2025intphys2}, this framework evaluates whether a model assigns higher surprise to events that contradict learned physical regularities.

Following prior work, we quantify surprise by measuring the discrepancy between the model’s predicted future observations and the actual observed future. We evaluate this framework across three environments: \textit{TwoRoom}, \textit{PushT}, and \textit{OGBench Cube}. For each environment, we introduce two types of perturbations. The first is a visual perturbation, where the color of an object changes abruptly during the trajectory. The second is a physical perturbation, where one or more objects are teleported to a random location, violating the expected physical continuity of the scene.  Fig.~\ref{fig:voe_surprise_lejepa} shows that LeWM consistently assigns higher surprise to frames containing physical violations compared to their unperturbed counterparts. We provide more details on VoE in App.~\ref{appendix:voe}.

\section{Conclusion}
We introduced LeWorldModel (LeWM), a stable end-to-end method for learning latent world models. LeWM is a Joint-Embedding Predictive Architecture in which an encoder maps image observations to a latent space and a predictor models temporal dynamics by forecasting future embeddings conditioned on actions. Across continuous control environments with raw pixel inputs, LeWM outperforms prior approaches in data efficiency, planning time, training time, and stability while remaining competitive in task performance. Training stability stems from explicitly encouraging latent embeddings toward an isotropic Gaussian distribution to prevent collapse, offering a scalable and principled alternative to existing work.
\paragraph{Limitations \& Future Work.} Several limitations point to future directions. Planning remains restricted to short horizons, motivating hierarchical world modeling for long-horizon reasoning. Our method also relies on offline datasets with sufficient coverage; in particular, low data diversity weakens SIGReg in simple, low-dimensional environments where matching a high-dimensional Gaussian prior is harder. Pre-training on large, diverse video datasets could provide stronger priors and reduce domain-specific data needs. Finally, dependence on action labels could be alleviated by inverse dynamics modeling.
\bibliography{ref}
\bibliographystyle{unsrtnat}


\appendix

\section{SIGReg}
\label{appendix:sigreg}
SIGReg proposes to match the distribution of embeddings towards the isotropic Gaussian target distribution. Achieving that match in high-dimension is gracefully done by combining two statistical components (i) Cramer-Wold theorem, and (ii) the univariate Epps-Pulley test-statistic. In short, SIGReg first produces $M$ unit-norm directions $\vu^{(m)}$ and projects the embeddings $\mZ$ onto them as
\begin{align}
    \vh^{(m)}&\triangleq \mZ\vu^{(m)},\vu^{(m)}\in\mathbb{S}^{D-1},\label{eq:h}
\end{align}
where the directions are sampled uniformly on the hypersphere. Then, SIGReg performs univariate distribution matching as
\begin{align}
    {\rm SIGReg}(\mZ)\triangleq \frac{1}{M}\sum_{m=1}^{M}T^{(m)},\tag{SIGReg}\label{eq:SIGreg}
\end{align}
with $T$ the univariate Epps-Pulley test-statistic
\begin{equation}
T^{(m)} = \int_{-\infty}^{\infty} w(t) \left|\phi_N(t; \vh^{(m)}) - \phi_0(t)\right|^2 dt,\tag{EP}\label{eq:EP}
\end{equation}
where the empirical characteristic function (ECF) is defined as $
\phi_N(t; \vh) = \frac{1}{N}\sum_{n=1}^N e^{it \vh_n}$, $w$ is a weighting function, e.g., $w(t) = e^{-\frac{t^2}{2\lambda^2}}$. Lastly, because the target is an isotropic Gaussian in $\mathbb{R}^D$, the univariate projection through $\vu^{(m)}$ makes the univariate target distribution $\phi_0$ the standard Gaussian $N(0,1)$. By Cramér–Wold, matching all 1D marginals implies matching the joint distribution, i.e., in the asymptotic limit over $M$ we have the following weak convergence result
\begin{align}
    {\rm SIGReg}(\mZ)\rightarrow 0 \iff \mathbb{P}_\mZ\rightarrow N(0,\mI).\tag{Cramer-Wold}
\end{align}
Practically, the integral in \eqref{eq:EP} employs a quadrature scheme, e.g., trapezoid with $T$ nodes uniformly distributed in $[0.2,4]$.

\section{Cross-Entropy Method}
\label{appendix:solver}

The \textit{Cross-Entropy Method} (CEM)~\cite{rubinstein2004cross} is a sampling-based (zero-order) optimization algorithm. Intuitively, CEM is an iterative sampling procedure that progressively refines a plan, defined as a sequence of actions, at each iteration.

At every iteration, the algorithm samples a pool of candidate plans from a distribution, 
typically a Gaussian (with initial parameters $\mu=\mathbf{0}$ and $\sigma=\mathbf{I}$). Next, each candidate plan is evaluated using the world model, and a cost is 
associated with it. The algorithm then selects the top $k$ plans with the lowest cost, referred 
to as \textit{elites}. These elites are used to compute statistics that update the parameters of 
the sampling distribution for the next iteration. Through this iterative process, the method 
explores the action space while gradually concentrating the sampling distribution around 
regions associated with lower costs. The final action plan is obtained from the mean of the 
sampling distribution at the last iteration.

However, in non-convex settings, there is no guarantee that the solution to which CEM converges 
is a global optimum. Furthermore, CEM suffers from the curse of dimensionality and becomes 
increasingly difficult to apply when the action space is large.

In our experiments, we use a CEM solver with $300$ sampled action sequences per iteration and 
perform $30$ optimization steps. At each step, the top $30$ candidates are selected as elites 
to update the sampling distribution. We provide the algorithm pseudo-code in Alg.~\ref{alg:cem}.

\setcounter{algorithm}{1}
\begin{algorithm}[h]
\caption{Cross-Entropy Method (CEM) for Action Sequence Optimization}
\label{alg:cem}
\begin{algorithmic}[1]
\Require World model $f$, planning horizon $H$, number of samples $N$, number of elites $K$, number of iterations $T$
\State Initialize sampling distribution parameters $\mu_0 = \mathbf{0}$, $\Sigma_0 = I$
\For{$t = 1$ to $T$}
    \State Sample $N$ candidate action sequences $\{a_{1:H}^{(i)}\}_{i=1}^{N} \sim \mathcal{N}(\mu_{t-1}, \Sigma_{t-1})$
    \For{$i = 1$ to $N$}
        \State Roll out $a_{1:H}^{(i)}$ in the world model $f$
        \State Compute cost $J^{(i)}$
    \EndFor
    \State Select the $K$ sequences with lowest cost (elites)
    \State Update distribution parameters using elite set:
    \State $\mu_t \leftarrow \frac{1}{K}\sum_{i \in \mathcal{E}} a_{1:H}^{(i)}$
    \State $\Sigma_t \leftarrow \text{Var}_{i \in \mathcal{E}}\left(a_{1:H}^{(i)}\right)$
\EndFor
\State \textbf{return} best action sequence found or first action of $\mu_T$
\end{algorithmic}
\end{algorithm}

\section{Baselines} \label{appendix:baselines}

\subsection{DINO-WM}

DINO world model (DINO-WM) focused on learning a predictor by leveraging DINOv2 frozen pre-trained representation to avoid collapse. Because not trained end-to-end, the loss simply is to minimize the predicted next-embedding with the ground trught next-state embedding produced by DINOv2. 

\begin{equation}
    \mathcal{L}_{\text{DINO-WM}} = \frac{1}{BT} \sum_i^B \sum_t^T \| \hat{\vz}^{(i)}_{t+1} - \vz^{(i)}_{t+1} \|_2^2
\end{equation}

We use the same setup as the original paper \citep{zhou2025dino-wm} (architecture, hyper-paremeters, etc..) 

\subsection{PLDM}

PLDM~\citep{sobal2025stresstesting} proposed a method for learning an end-to-end joint-embedding predictive architecture (JEPA). To avoid collapse, their approach takes inspiration from the variance-invariance-covariance regularization (VICReg, \cite{bardes2022vicreg}) with extra terms to take into account the temporality of the next state prediction. The PLDM objective is the following:

\begin{equation}
    \mathcal{L}_{\text{PLDM}} =  \mathcal{L}_{\text{pred}} +  \alpha \mathcal{L}_{\text{var}} +  \beta \mathcal{L}_{\text{cov}} +  \gamma \mathcal{L}_{\text{time-sim}} + \zeta \mathcal{L}_{\text{time-var}} + \nu \mathcal{L}_{\text{time-cov}} + \mu
    \mathcal{L}_{\text{IDM}}
\end{equation}

where,

\begin{equation*}
    \mathcal{L}_{\text{pred}} = \frac{1}{BT} \sum_i^B \sum_t^T \| \hat{\vz}^{(i)}_{t+1} - \vz^{(i)}_{t+1}\|_2^2
\end{equation*}

\begin{equation*}
    \mathcal{L}_{\text{var}} =  \frac{1}{TD} \sum_t^T \sum_d^D \max\left(0,  1-\sqrt{\text{Var}(\vz^{(:)}_{t,d})}+\epsilon\right)
\end{equation*}

\begin{equation*}
    \mathcal{L}_{\text{cov}} =  \frac{1}{T} \sum_t^T \frac{1}{D}\sum_{i\neq j}^D \left[\text{Cov}(\mZ_t)\right]_{ij}
\end{equation*}

\begin{equation*}
    \mathcal{L}_{\text{time-sim}} = \frac{1}{BT} \sum_i^B \sum_t^T \| \vz^{(i)}_t - \vz^{(i)}_{t+1}\|_2^2
\end{equation*}

\begin{equation*}
    \mathcal{L}_{\text{time-var}} =  \frac{1}{BD} \sum_i^B \sum_d^D \max\left(0,  1-\sqrt{\text{Var}(\vz^{(i)}_{:,d})}+\epsilon\right)
\end{equation*}

\begin{equation*}
    \mathcal{L}_{\text{time-cov}} =  \frac{1}{B} \sum_b^B \frac{1}{D}\sum_{i\neq j}^D \left[\text{Cov}(\mZ)\right]_{ij}
\end{equation*}

\begin{equation*}
    \mathcal{L}_{\text{IDM}} = \frac{1}{BT} \sum_i^B \sum_t^T \| \hat{\va}^{(i)}_t - \va^{(i)}_t\|_2^2
\end{equation*}
with $\vz^{(i)}_{t} \in \R^{D}$ correspond to step $t \in [T]$ of trajectory $i \in [B]$ and $T$ is trajectory length and $B$ the batch size, and $\mZ_t \in \mathbb{R}^{B \times D}$ denote the matrix whose $i$-th row is $\vz_t^{(i)}$, i.e.,
\[
\mZ_t =
\begin{bmatrix}
(\vz_t^{(1)})^\top \\
\vdots \\
(\vz_t^{(B)})^\top
\end{bmatrix},
\]
Let $\bar{\mZ}_t$ be the row-centered version of $\mZ_t$:
\[
\bar{\mZ}_t = \mZ_t - \frac{1}{B}\mathbf{1}\mathbf{1}^\top \mZ_t .
\]
Then, for each time step $t$ and feature dimension $d$, the variance across the batch is
\[
\mathrm{Var}(\vz^{(:)}_{t,d})
=
\frac{1}{B-1}\sum_{i=1}^B
\left(
z^{(i)}_{t,d} - \frac{1}{B}\sum_{i'=1}^B z^{(i')}_{t,d}
\right)^2 ,
\]
and the covariance matrix across feature dimensions is
\[
\mathrm{Cov}(\mZ_t)
=
\frac{1}{B-1}\bar{\mZ}_t^\top \bar{\mZ}_t
\in \mathbb{R}^{D \times D}.
\]

Similarly, for the temporal regularization, let $\mZ^{(i)} \in \mathbb{R}^{T \times D}$ denote the matrix whose $t$-th row is $\vz_t^{(i)}$, and let $\bar{\mZ}^{(i)}$ be its row-centered version:
\[
\bar{\mZ}^{(i)} = \mZ^{(i)} - \frac{1}{T}\mathbf{1}\mathbf{1}^\top \mZ^{(i)} .
\]
Then the variance across time is
\[
\mathrm{Var}(\vz^{(i)}_{:,d})
=
\frac{1}{T-1}\sum_{t=1}^T
\left(
z^{(i)}_{t,d} - \frac{1}{T}\sum_{t'=1}^T z^{(i)}_{t',d}
\right)^2 ,
\]
and the temporal covariance matrix is
\[
\mathrm{Cov}(\mZ^{(i)})
=
\frac{1}{T-1}(\bar{\mZ}^{(i)})^\top \bar{\mZ}^{(i)}
\in \mathbb{R}^{D \times D}.
\]

$\hat{\vz}^{(i)}_{t} \in \R^{d}$ is the predicted embedding at step $t$ for traj $i$ using the predictor. $\va^{(i)}_{t} \in \R^{A}$ is the action associated to step $t$ and $\hat{\va}^{(i)}_{t} \in \R^{A}$ is the predicted action for the inverse dynamic model (IDM) $\text{idm}(\vz_t, \vz_{t+1})$.

We select PLDM hyperparameters via a grid search over the loss coefficients. Since the overall objective includes six tunable weights ($\alpha$, $\beta$, $\gamma$, $\zeta$, $\nu$, $\mu$), an exhaustive search over all combinations is not tractable $(\mathcal{O}(n^6))$. Moreover, the original PLDM study reports coefficients that were extensively tuned per environment and dataset, which limits their transferability. We start from the set of hyperparameters from the config provided in their open-source codebase. We motivate this choice by mentioning that no mention of the time-var and time-cov regularization term are mentionned in the original paper. We then perform a grid search for each initial loss coefficient over 256 configurations on Push-T and keep the one performing the best on a held-out set. We report the best hyperparameters found in Table \ref{tab:pldm_init_coeffs}. We kept these coefficients fixed for all training.

\begin{table}[h]
\centering
\begin{tabular}{l c}
\toprule
\textbf{Loss coefficient} & \textbf{Initial value} \\

\toprule
$\alpha$ & 18.0 \\
$\beta$  & 12   \\
$\gamma$ & 0.2  \\
$\zeta$  & 0.7  \\
$\nu$    & 0.0  \\
$\mu$    & 0.0  \\
\bottomrule
\end{tabular}
\vspace{2mm}
\caption{Best coefficient found from grid search.}
\label{tab:pldm_init_coeffs}
\end{table}

\subsection{GC-RL}

To evaluate downstream control, we use goal-conditioned reinforcement learning (GC-RL) with offline training. In particular, we consider goal-conditioned variants of Implicit Q-Learning (IQL) and Implicit Value Learning (IVL). In both cases, observations and goals are encoded using DINOv2 patch embeddings, and policies are trained from offline datasets. Training proceeds in two phases: first learning a value function (and optionally a Q-function), followed by policy extraction via advantage-weighted regression.

\paragraph{GCIQL}
Implicit Q-Learning (IQL) \cite{kostrikov2021offline} is an offline reinforcement learning algorithm that avoids querying out-of-distribution actions by learning a value function via expectile regression. In the goal-conditioned setting, the algorithm learns both a Q-function $Q_\psi(s_t, a_t, g)$ and a value function $V_\theta(s_t, g)$ conditioned on a goal $g$.

The Q-function is trained with Bellman regression, bootstrapping from a target value network $V_{\bar{\theta}}$:

\[
\mathcal{L}_{Q} =
\mathbb{E}_{(s_t, a_t, s_{t+1}, g) \sim \mathcal{D}}
\left[
\left(
Q_\psi(s_t, a_t, g) -
\left(
r(s_t, g) + \gamma m_t V_{\bar{\theta}}(s_{t+1}, g)
\right)
\right)^2
\right],
\]

where $m_t = 0$ if $s_t = g$ (terminal transition) and $m_t = 1$ otherwise.

The value network is trained using expectile regression against targets from the target Q-network $Q_{\bar{\psi}}$:

\[
\mathcal{L}_{V} =
\mathbb{E}_{(s_t, a_t, g) \sim \mathcal{D}}
\left[
L_\tau^2
\left(
Q_{\bar{\psi}}(s_t, a_t, g) - V_\theta(s_t, g)
\right)
\right],
\]

where the expectile loss is defined as

\[
L_\tau^2(u) = |\tau - \mathbbm{1}(u < 0)| u^2 .
\]

The total critic loss is given by

\[
\mathcal{L}_{\text{critic}} = \mathcal{L}_Q + \mathcal{L}_V .
\]

\paragraph{GCIVL}
Implicit Value Learning (IVL) \cite{park2025ogbench} simplifies IQL by removing the Q-function and learning the value function directly through bootstrapped targets. The value network $V_\theta(s_t, g)$ is trained via expectile regression against a target network $V_{\bar{\theta}}$:

\[
\mathcal{L}_{V} =
\mathbb{E}_{(s_t, s_{t+1}, g) \sim \mathcal{D}}
\left[
L_\tau^2
\left(
r(s_t, g) + \gamma V_{\bar{\theta}}(s_{t+1}, g)
-
V_\theta(s_t, g)
\right)
\right].
\]

As in IQL, $L_\tau^2$ denotes the asymmetric expectile loss and $\gamma$ is the discount factor.

\paragraph{Policy extraction.}

For both GCIQL and GCIVL, the policy $\pi_\theta(s_t, g)$ is trained via advantage-weighted regression (AWR). The policy objective is

\[
\mathcal{L}_{\pi} =
\mathbb{E}_{(s_t, a_t, g) \sim \mathcal{D}}
\left[
\exp\left(\beta A(s_t, a_t, g)\right)
\|
\pi_\theta(s_t, g) - a_t
\|_2^2
\right],
\]

where the advantage is computed as

\[
A(s_t, a_t, g) =
r(s_t, g) + \gamma V(s_{t+1}, g) - V(s_t, g),
\]

and $\beta$ is an inverse temperature parameter controlling the strength of advantage weighting.

\subsection{GCBC}

As a simple imitation learning baseline, we consider Goal-Conditioned Behavioral Cloning (GCBC) \cite{ghosh2019learning}. GCBC trains a goal-conditioned policy $\pi_\theta(s_t, g)$ to reproduce expert actions given the current observation $s_t$ and a goal observation $g$. In our implementation, both observations and goals are encoded using DINOv2 patch embeddings before being provided to the policy network.

The policy is trained via supervised learning on an offline dataset $\mathcal{D}$ of state-action-goal tuples. Specifically, the objective minimizes the mean squared error between the predicted action and the action taken in the dataset:

\[
\mathcal{L}_{\text{GCBC}} =
\mathbb{E}_{(s_t, a_t, g) \sim \mathcal{D}}
\left[
\|
\pi_\theta(s_t, g) - a_t
\|_2^2
\right],
\]

where $s_t$ denotes the observation embedding, $g$ the goal embedding, and $a_t$ the corresponding expert action.

\section{Implementation details}
\label{appendix:details}
We apply a frame-skip of 5, grouping consecutive actions between frames into a single action block. This choice enables computationally efficient longer-horizon predictions while maintaining informative temporal transitions. 
We use a batch size of 128 with sub-trajectories of size 4 corresponding to 4 frames and 4 blocks of 5 actions. Each frame is $224\times224$ pixels.

\begin{figure}[t]
\centering
\captionsetup{labelformat=empty}
\caption{\textbf{Algorithm \ref*{alg:train-alg}.} Pseudo-code for the training procedure of LeWorldModel. Pixel observations are encoded into latent embeddings, and a predictor estimates the dynamics by predicting the next-step embedding conditioned on actions. The model is optimized end-to-end using a next-embedding prediction loss together with a step-wise SIGReg regularization term to prevent representation collapse.}
\addtocounter{figure}{-1}
\refstepcounter{algorithm}
\label{alg:train-alg}
\begin{tcolorbox}[
  colback=codebg,
  colframe=codebg,
  arc=7pt,
  outer arc=7pt,
  boxrule=0pt,
  left=3pt, right=3pt, top=3pt, bottom=3pt,
  width=0.9\linewidth
]
\begin{lstlisting}[style=pythonstyle]
def LeWorldModel(obs,actions,lambd=0.1):
    """
    obs: (B, T, C, H, W) raw pixels sequence
    actions: (B, T, A) action sequence
    lambd: (float) SIGReg loss weight
    """
    
    emb = encoder(obs) # (B, T, D)
    next_emb = predictor(emb,actions) #(B, T, D)
    
    # -- LeWorldModel training loss
    
    # next-embedding prediction loss
    pred_loss = F.mse_loss(emb[:, 1:] - next_emb[:, :-1])
    
    # step-wise sigreg (anti-collapse)
    sigreg_loss = mean(SIGReg(emb.transpose(0, 1))
    
    return pred_loss + lambd * sigreg_loss
\end{lstlisting}
\end{tcolorbox}
\end{figure}

\paragraph{Encoder Architecture.}
The encoder is a Vision Transformer Tiny (ViT-Tiny) model from the Hugging Face library, using a patch size of 14.

\paragraph{Predictor Architecture.}
The predictor is implemented as a ViT-S backbone with learned positional embeddings and causal masking over the observation history. The history length is set to 3 for the \textit{PushT} and \textit{OGBench-Cube} environments, and to 1 for \textit{TwoRoom}. During planning, the predictor is used autoregressively to generate rollouts of future latent states.

\paragraph{Decoder (Visualization Only).}
For visualization, we decode the \verb|[CLS]| token embedding (192 dim) from the last encoder layer into an image using a lightweight transformer decoder. The \verb|[CLS]| representation is first projected to a hidden dimension and used as the key and value in cross-attention. A fixed set of learnable query tokens, one for each patch of the target image, interacts with this global representation through several cross-attention layers with residual MLP blocks. For an image of size $224\times224$ with patch size $16$, this corresponds to $P=(224/16)^2=196$ learnable query tokens. The resulting patch embeddings are then linearly projected to $16 \times 16 \times 3$ pixel patches and rearranged to produce a $224 \times 224$ RGB image. This decoder is used only as a diagnostic tool to visualize what visual information is retained in the \verb|[CLS]| representation.

\paragraph{Planning solver.}
For planning, we use the Cross-Entropy Method (CEM). At each planning step, CEM samples 300 candidate action sequences and optimizes them for a maximum of 30 iterations in \textit{PushT} and 10 iterations in the other environments. At each iteration, the top 30 trajectories are retained to update the sampling distribution, and the initial sampling variance is set to 1. The planning horizon is set to 5 steps, which corresponds to 25 environment timesteps due to the use of a frame skip of 5. We employ a receding-horizon Model Predictive Control (MPC) scheme with a horizon of 5, meaning that the entire optimized action sequence is executed before replanning. This configuration follows the setup used in \cite{zhou2025dino-wm}.

\section{Environment \& Dataset} \label{appendix:envs}

\begin{enumerate}[a)]

\item \textbf{TwoRoom} is a simple continuous 2D navigation task introduced by \citet{sobal2025stresstesting}. The environment consists of two rooms separated by a wall with a single door connecting them. The agent (represented as a red dot) must navigate from a random starting position in one room to a randomly sampled target location in the other room, which requires passing through the door. We collect 10,000 episodes with an average trajectory length of 92 steps. The data are generated using a simple noisy heuristic policy that first directs the agent toward the door along a straight-line path and then toward the target location once the agent has crossed into the other room. Each world model is trained on this dataset for 10 epochs.

\item \textbf{PushT} is a continuous 2D manipulation task in which an agent (represented as a blue dot) must push a T-shaped block to match a target configuration, with interactions restricted to pushing actions. We follow the same setup and dataset as \citet{zhou2025dino-wm}, which contains 20,000 expert episodes with an average length of 196 steps. However, we train each world model for only 10 epochs. Empirically, we observe that 10 epochs are sufficient to reach the best performance, matching the results reported in the DINO-WM paper.

\item \textbf{OGBench-Cube} is a continuous 3D robotic manipulation task in which a robotic arm with an end-effector must pick up a cube and place it at a target location. Originally introduced by \citet{park2025ogbench}, we consider only the single-cube variant. We collect 10,000 episodes, each consisting of 200 steps. The data are generated using the data-collection heuristic provided in the benchmark library. Each world model is trained on this dataset for 10 epochs.

\item \textbf{Reacher} is a continuous control environment from the DeepMind Control Suite~\cite{tassa2018deepmind}. The task consists of controlling a two-joint robotic arm to reach a target location in a 2D plane. Following the setup used in DINO-WM, we consider the variant where success is defined by the perfect alignment of the arm joints with the target configuration required to reach the goal position. We train each world model for 10 epochs on a dataset of 10,000 episodes, each with 200 steps. The data are collected using a Soft Actor-Critic policy.

\end{enumerate}

\section{Evaluation Details}

\begin{figure}[t]
    \centering
    \includegraphics[width=\linewidth]{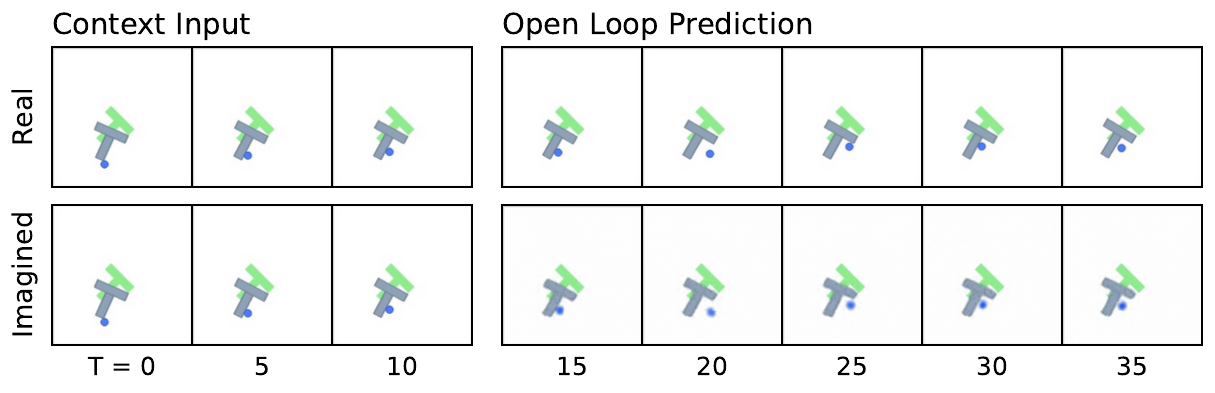}\\
    \includegraphics[width=\linewidth]{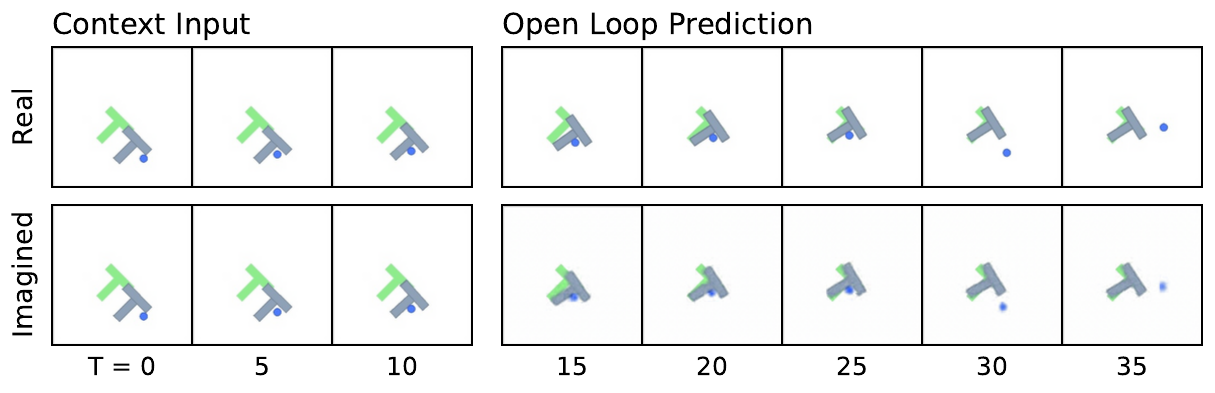}\\
    \includegraphics[width=\linewidth]{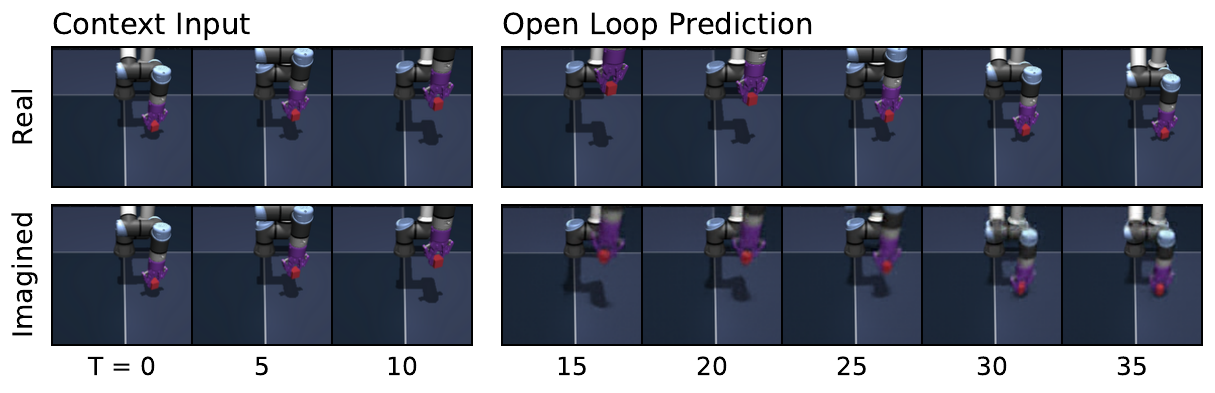}
    \caption{\small \textbf{Additional predictor rollouts on PushT (top) and OGBench-Cube (bottom).} Same setup as Fig.~\ref{fig:rollout}: three context frames are encoded into latent representations, and the predictor autoregressively generates future latent states conditioned on the action sequence. All predictions are decoded using a decoder not used during training. On PushT, the imagined trajectory closely tracks the real one, accurately capturing both agent and block motion. On OGBench-Cube, the model preserves the overall scene layout and cube displacement but loses finer details such as end-effector orientation at longer horizons, consistent with the lower probing accuracy on rotational quantities reported in Tab.~\ref{tab:probe-cube}.}
    \label{fig:rollout-2}
\end{figure}

\subsection{Control} \label{appendix:control_details}

We evaluate LeWM on goal-conditioned control tasks in the three environments introduced previously. Control performance is measured using two parameters: the evaluation budget and the distance to the goal. The evaluation budget corresponds to the maximum number of actions the agent is allowed to execute in the environment. The goal distance determines how far in the future the goal state is sampled relative to the initial state.
During evaluation, trajectories are sampled from the offline dataset. The initial state is chosen by randomly sampling a state from a trajectory in the dataset, while the goal state corresponds to a state occurring several timesteps later in the same trajectory. This ensures that the goal is reachable and consistent with the dataset dynamics.
In \textit{TwoRoom}, the evaluation budget is set to 50 steps, and the goal state is sampled 25 timesteps in the future. In \textit{PushT}, the evaluation budget is 50 steps and the goal is sampled 25 timesteps in the future. In \textit{OGBench-Cube} and \textit{Reacher}, the evaluation budget is 50 steps, and the goal is sampled 25 timesteps in the future.

\subsection{Probing} \label{appendix:probing}

\begin{figure}[h]
    \centering
    \includegraphics[width=\linewidth]{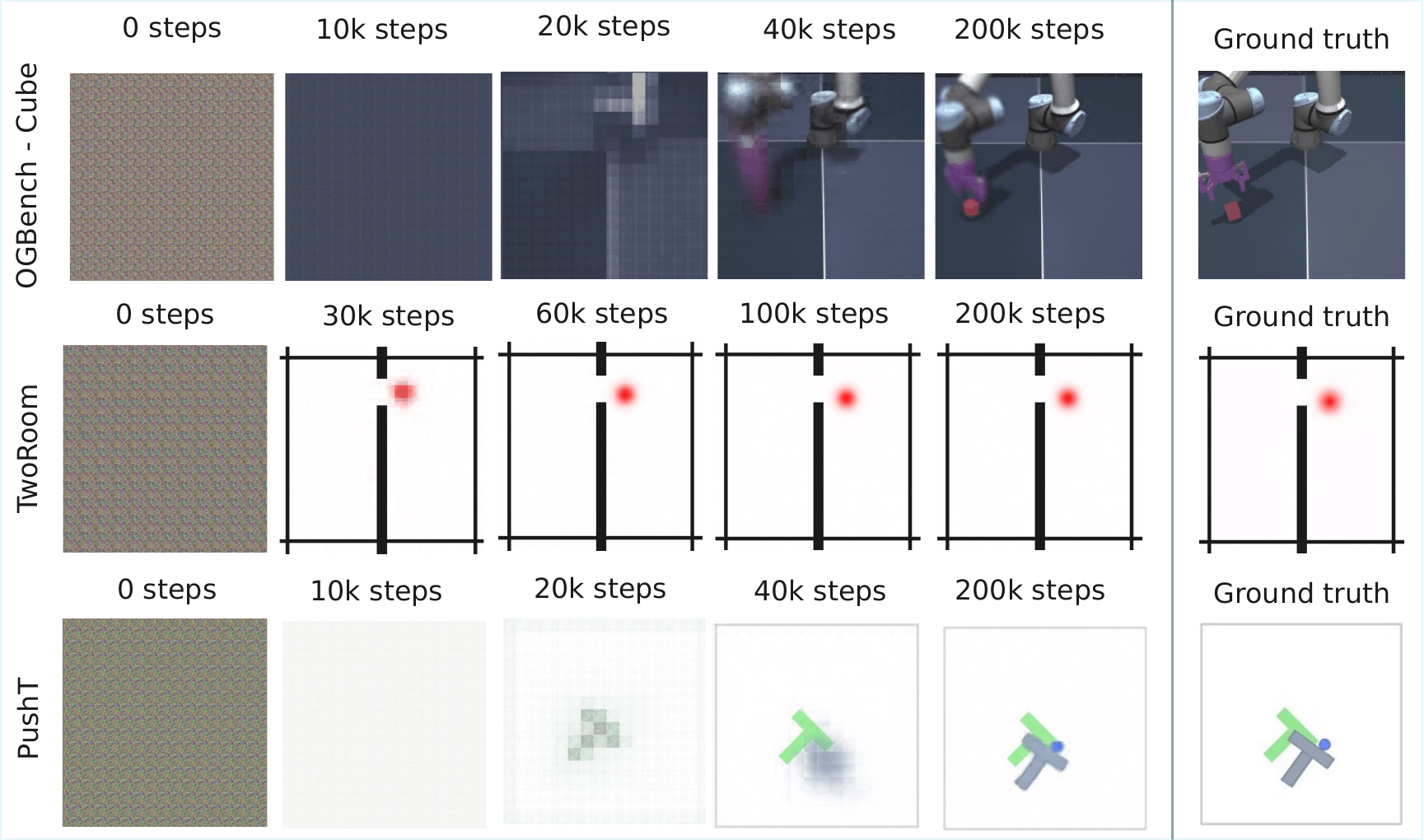}
    \caption{\small \textbf{Decoder visualization during training.} As training progresses, the latent representation increasingly captures the information required to reconstruct the visual scene, even though no reconstruction loss is used during training. Early in training, the decoded images correspond to slow features, a phenomenon previously reported~\cite{sobal2022jointembeddingpredictivearchitectures}.}
    \label{fig:decoder-vis}
\end{figure}

We use probing to analyze the information contained in the learned latent representations across the three environments. Specifically, we train both linear and non-linear probes to predict physical quantities from the latent embeddings. Linear probes evaluate whether the information is linearly accessible in the latent space, while non-linear probes assess whether the information is present but potentially entangled.

For each probe, we report the mean squared error (MSE) and the Pearson correlation coefficient between the predicted and ground-truth quantities.

The probed variables differ across environments. In \textit{TwoRoom}, we probe the 2D position of the agent (Tab.~\ref{tab:probe-tworoom}). In \textit{PushT}, we probe both the state of the agent and the state of the block (Tab.~\ref{tab:probe-pusht}). In \textit{OGBench-Cube}, we probe the position of the cube and the position of the robot end-effector (Tab.~\ref{tab:probe-cube}).

\begin{table}[h]
\centering
\caption{\small \textbf{Physical Latent Probing results on TwoRoom.} Although LeWM underperforms PLDM in downstream planning on this environment, it matches or outperforms PLDM across all probing metrics, and both methods substantially outperform DINO-WM on the linear probe. This suggests that the learned latent space captures the underlying physical state equally well and that the planning gap is not due to a less informative representation but rather to other factors such as the dynamics model or the planning procedure itself.}
\begin{tabular}{l c c c c}
\toprule
& \multicolumn{4}{c}{\textbf{Agent Position}}\\
\cmidrule(lr){2-5}
& \multicolumn{2}{c}{Linear} & \multicolumn{2}{c}{MLP}\\
\cmidrule(lr){2-3} \cmidrule(lr){4-5}
\textbf{Model} & MSE $\downarrow$ & r $\uparrow$ & MSE $\downarrow$ & r $\uparrow$\\
\toprule
DINO-WM & $0.488 \pm 0.451$ & $0.824$ & $0.000 \pm 0.000$ & $0.999$\\
PLDM & $0.008 \pm 0.041$ & $0.996$ & $0.000 \pm 0.000$ & $1.000$\\
\midrule
LeWM & $0.008 \pm 0.018$ & $0.996$ & $0.000 \pm 0.000$ & $1.000$\\ 
\bottomrule
\end{tabular}
\vspace{2mm}
\label{tab:probe-tworoom}
\end{table}

\begin{table}[h]
\centering
\caption{\small \textbf{Physical latent probing results on OGBench-Cube.} LeWM matches or outperforms PLDM on most properties and achieves the best results on positional quantities such as block position and end-effector position. DINO-WM retains a clear advantage on dynamic and rotational properties (joint velocity, end-effector yaw), likely because such quantities benefit from the richer visual priors learned during large-scale pretraining. All three methods struggle to recover block orientation (quaternion and yaw), suggesting that fine-grained rotational information remains difficult to encode in compact latent spaces
regardless of the training strategy.}
\resizebox{\textwidth}{!}{%
\begin{tabular}{l l c c c c}
\toprule
& & \multicolumn{2}{c}{\textbf{Linear}} & \multicolumn{2}{c}{\textbf{MLP}}\\
\cmidrule(lr){3-4} \cmidrule(lr){5-6}
\textbf{Property} & \textbf{Model} & MSE $\downarrow$ & r $\uparrow$ & MSE $\downarrow$ & r $\uparrow$\\
\toprule
\multirow{3}{*}{Joint Position}
  & DINO-WM & $0.960 \pm 1.150$ & $0.808$ & $0.200 \pm 0.967$ & $0.870$\\
  & PLDM    & $0.372 \pm 1.172$ & $0.695$ & $0.340 \pm 1.164$ & $0.728$\\
  & LeWM    & $0.352 \pm 1.173$ & $0.706$ & $0.330 \pm 1.157$ & $0.742$\\
\midrule
\multirow{3}{*}{Joint Velocity}
  & DINO-WM & $0.792 \pm 0.748$ & $0.763$ & $0.263 \pm 0.683$ & $0.852$\\
  & PLDM    & $1.016 \pm 0.905$ & $0.115$ & $0.661 \pm 0.830$ & $0.536$\\
  & LeWM    & $1.021 \pm 0.902$ & $0.095$ & $0.818 \pm 0.899$ & $0.386$\\
\midrule
\multirow{3}{*}{End-Effector Position}
  & DINO-WM & $0.024 \pm 0.010$ & $0.996$ & $0.004 \pm 0.003$ & $0.999$\\
  & PLDM    & $0.052 \pm 0.073$ & $0.974$ & $0.013 \pm 0.029$ & $0.993$\\
  & LeWM    & $0.018 \pm 0.025$ & $0.991$ & $0.003 \pm 0.004$ & $0.998$\\
\midrule
\multirow{3}{*}{End-Effector Yaw}
  & DINO-WM & $3.317 \pm 1.016$ & $0.828$ & $0.167 \pm 0.168$ & $0.917$\\
  & PLDM    & $0.996 \pm 0.165$ & $0.056$ & $0.985 \pm 0.207$ & $0.117$\\
  & LeWM    & $0.980 \pm 0.295$ & $0.124$ & $0.952 \pm 0.369$ & $0.213$\\
\midrule
\multirow{3}{*}{Gripper}
  & DINO-WM & $0.114 \pm 0.095$ & $0.943$ & $0.038 \pm 0.060$ & $0.982$\\
  & PLDM    & $0.234 \pm 0.169$ & $0.876$ & $0.066 \pm 0.111$ & $0.967$\\
  & LeWM    & $0.121 \pm 0.111$ & $0.938$ & $0.048 \pm 0.079$ & $0.976$\\
\midrule
\multirow{3}{*}{Block Position}
  & DINO-WM & $0.085 \pm 0.029$ & $0.991$ & $0.007 \pm 0.007$ & $0.998$\\
  & PLDM    & $0.031 \pm 0.023$ & $0.985$ & $0.003 \pm 0.004$ & $0.999$\\
  & LeWM    & $0.007 \pm 0.010$ & $0.997$ & $0.002 \pm 0.003$ & $0.999$\\
\midrule
\multirow{3}{*}{Block Quaternion}
  & DINO-WM & $1.596 \pm 10.457$ & $0.257$ & $0.769 \pm 8.046$ & $0.411$\\
  & PLDM    & $1.021 \pm 12.600$ & $0.066$ & $0.989 \pm 12.140$ & $0.218$\\
  & LeWM    & $1.019 \pm 12.596$ & $0.087$ & $0.963 \pm 11.450$ & $0.224$\\
\midrule
\multirow{3}{*}{Block Yaw}
  & DINO-WM & $4.223 \pm 2.530$ & $0.176$ & $0.916 \pm 0.278$ & $0.304$\\
  & PLDM    & $0.996 \pm 0.088$ & $0.061$ & $0.989 \pm 0.140$ & $0.106$\\
  & LeWM    & $0.996 \pm 0.094$ & $0.062$ & $0.973 \pm 0.199$ & $0.164$\\
\midrule
\multirow{3}{*}{\textbf{Overall}}
  & DINO-WM & $1.162 \pm 1.579$ & $0.725$ & $0.290 \pm 1.202$ & $0.799$\\
  & PLDM    & $0.611 \pm 1.875$ & $0.464$ & $0.503 \pm 1.809$ & $0.600$\\
  & LeWM    & $0.592 \pm 1.874$ & $0.477$ & $0.525 \pm 1.714$ & $0.584$\\
\bottomrule
\end{tabular}%
}
\label{tab:probe-cube}
\end{table}

\subsection{Violation-of-expectation}\label{appendix:voe}

\begin{figure}
    \centering
    \includegraphics[width=\linewidth]{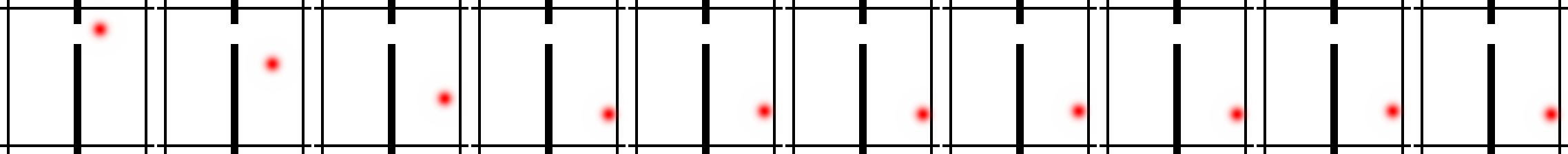}
    \includegraphics[width=\linewidth]{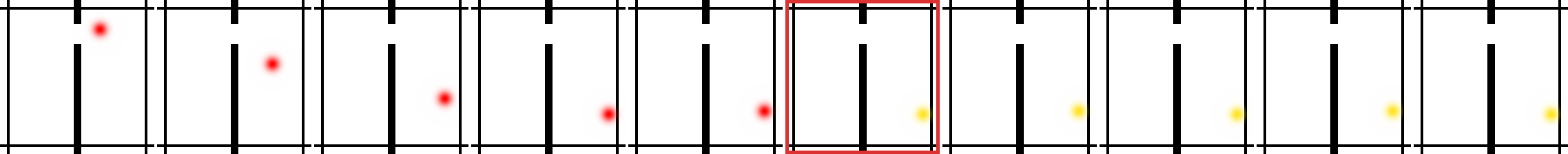}
    \includegraphics[width=\linewidth]{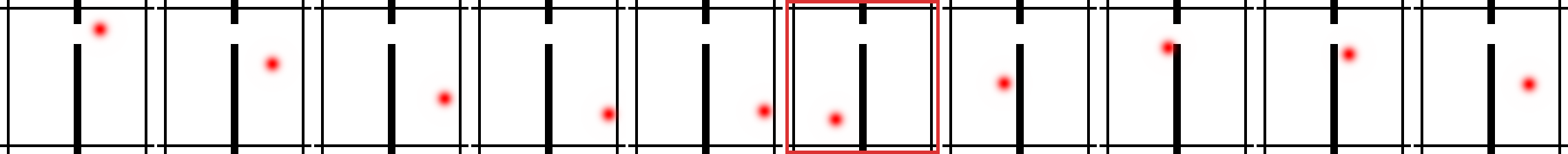}
    \includegraphics[width=\linewidth]{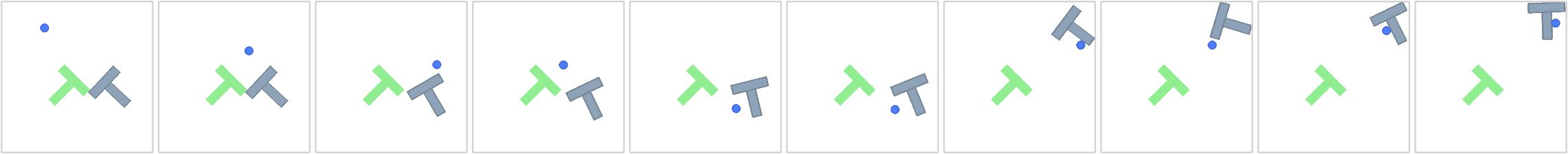}
    \includegraphics[width=\linewidth]{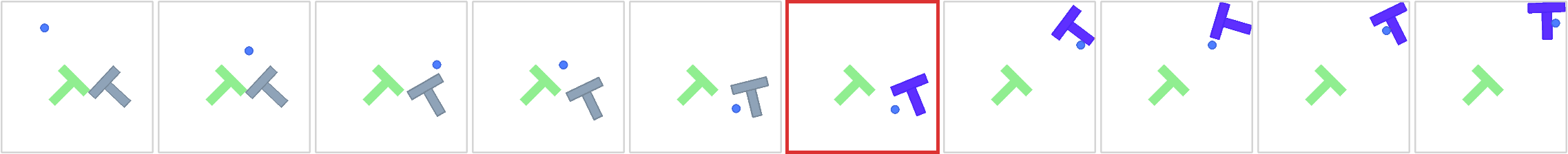}
    \includegraphics[width=\linewidth]{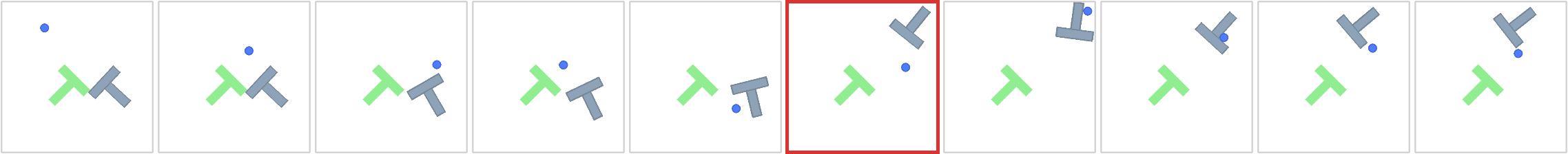}
    \includegraphics[width=\linewidth]{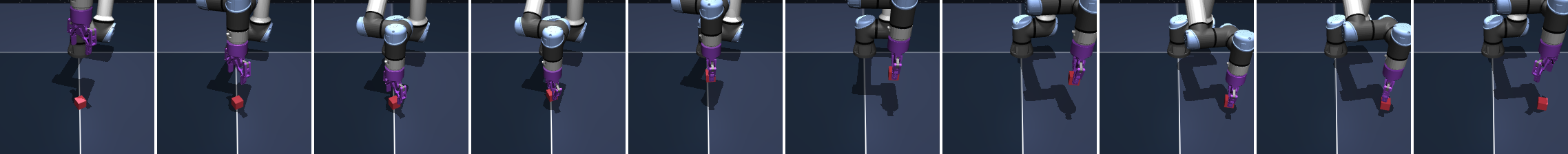}
    \includegraphics[width=\linewidth]{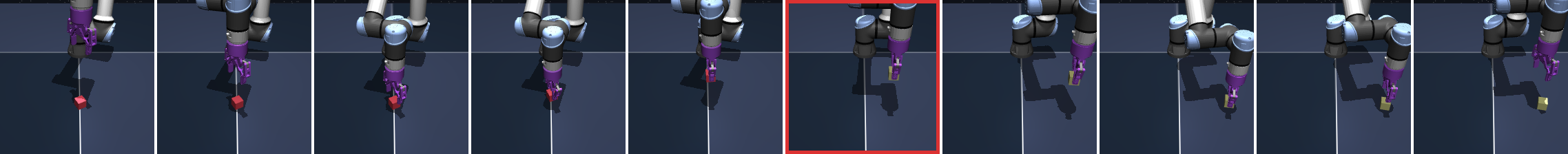}
    \includegraphics[width=\linewidth]{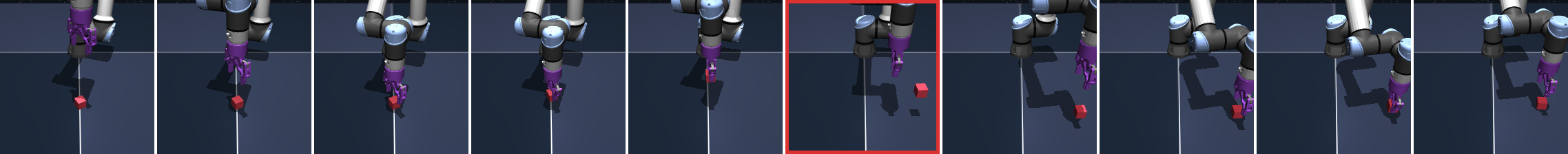}
    \caption{\small \textbf{Example of trajectories used for the Violation of Expectation experiments} (Sec.~\ref{subsec:voe}). For each environment, the first row corresponds to the unperturbed trajectory, the second row corresponds to a trajectory where a visual perturbation occurs and the third row displays trajectories where the state of the system is randomly reset in the middle of the trajectory. The frame where the perturbation occurs is highlighted in red.}
    \label{fig:voe_frames}
\end{figure}

We evaluate physical understanding using the violation-of-expectation (VoE) framework across three environments. In each environment, we generate three types of trajectories: an unperturbed reference trajectory, a trajectory containing a visual perturbation, and a trajectory containing a physical perturbation. Visual perturbations correspond to abrupt color changes of an object, while physical perturbations correspond to teleporting objects to random positions, thereby violating physical continuity. Examples of trajectories are shown in Figure~\ref{fig:voe_frames}.

\begin{figure}
    \centering
    \includegraphics[width=0.32\linewidth]{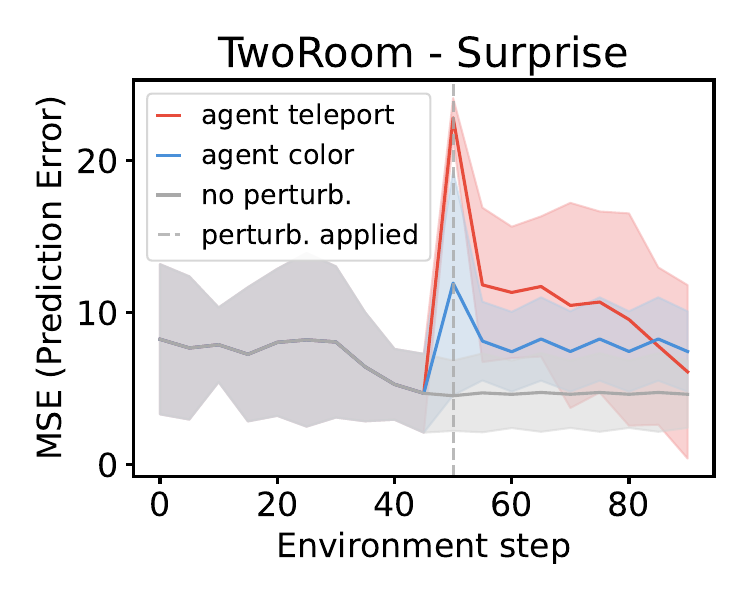}
    \includegraphics[width=0.32\linewidth]{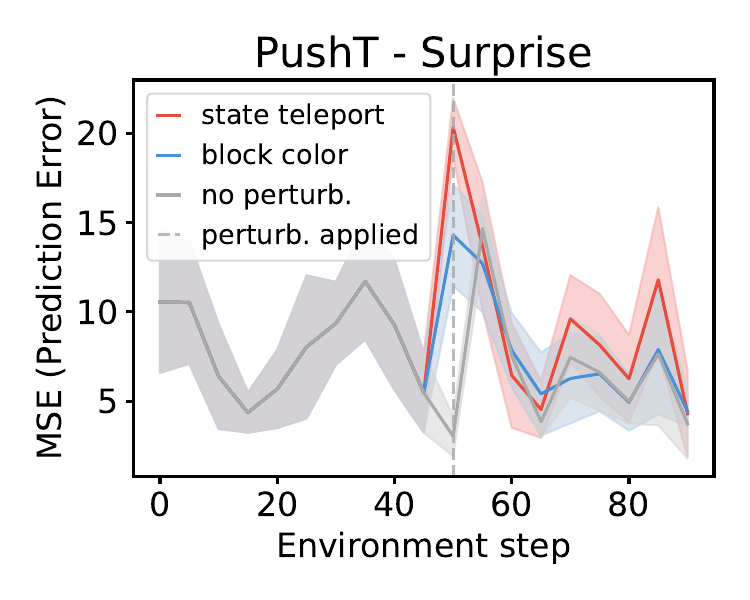}
    \includegraphics[width=0.32\linewidth]{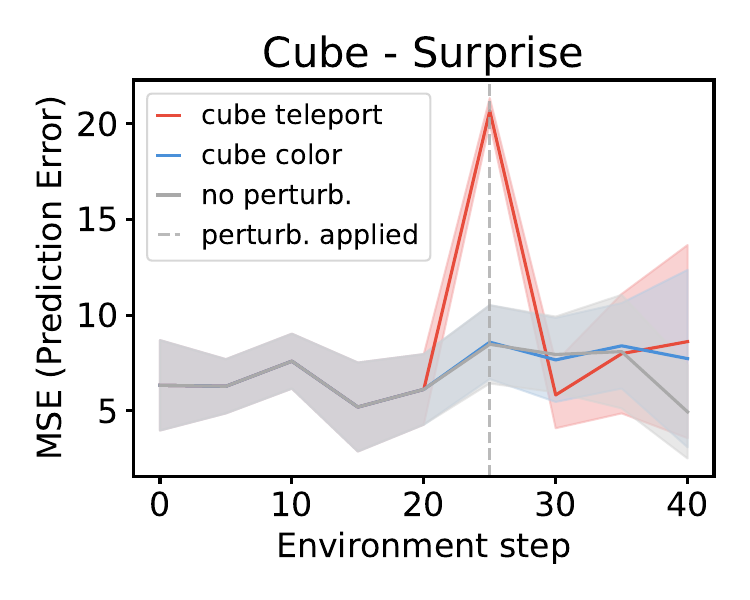}
    \caption{\textbf{Violation-of-expectation evaluation with PLDM.} From left to right: \textit{TwoRoom}, \textit{PushT}, and \textit{OGBench-Cube}. Surprise is plotted over time for unperturbed, visually perturbed, and physically perturbed trajectories. In \textit{TwoRoom} and \textit{PushT}, the model assigns significantly higher surprise to both visual and physical perturbations. In \textit{OGBench-Cube}, the increase in surprise is weaker and not consistently significant.}
    \label{fig:voe_surprise_pldm}
\end{figure}

\begin{figure}
    \centering
    \includegraphics[width=0.3\linewidth]{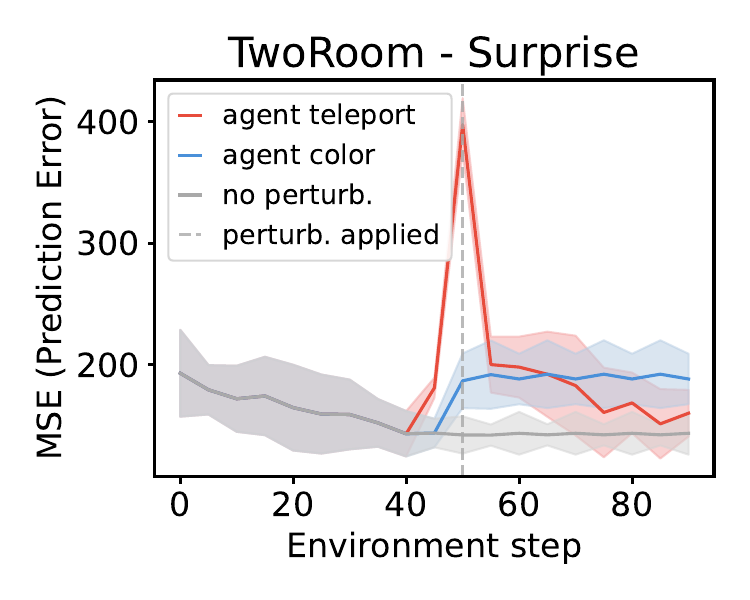}
    \includegraphics[width=0.3\linewidth]{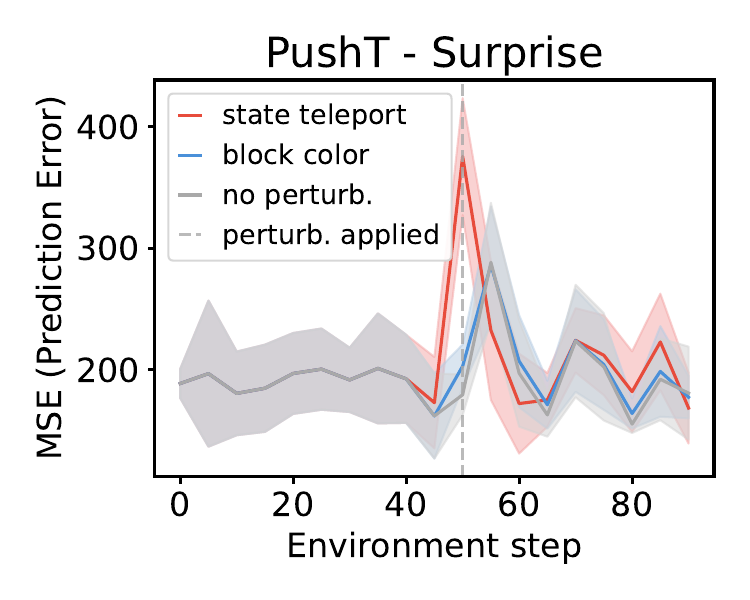}
    \includegraphics[width=0.3\linewidth]{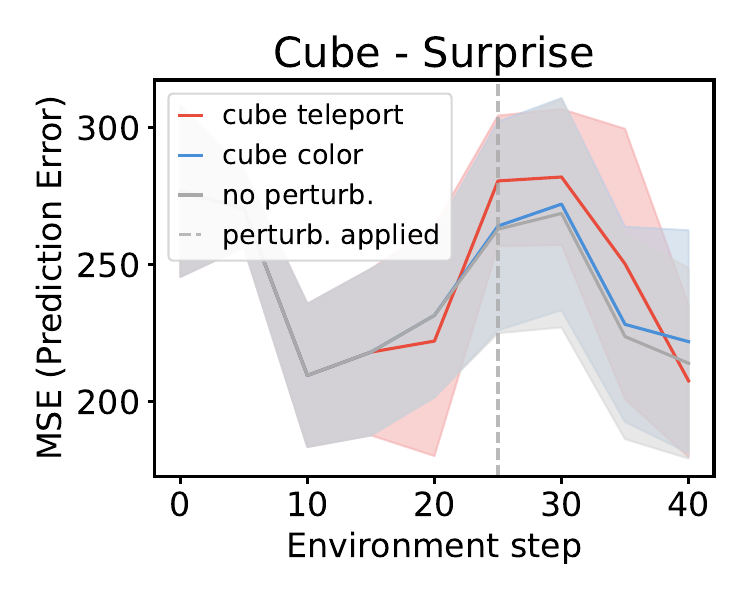}
    \caption{\textbf{Violation-of-expectation evaluation with DINO-WM.} From left to right: \textit{TwoRoom}, \textit{PushT}, and \textit{OGBench-Cube}. Surprise is plotted over time for unperturbed, visually perturbed, and physically perturbed trajectories. While the model detects both perturbations in \textit{TwoRoom} and \textit{PushT}, surprise does not increase significantly for either perturbation in \textit{OGBench-Cube}.}
    \label{fig:voe_surprise_dinowm}
\end{figure}

\paragraph{TwoRoom.}
In the \textit{TwoRoom} environment, the agent is controlled by an expert policy that navigates toward a goal position. We generate three trajectories: (1) an unperturbed trajectory, (2) a trajectory where the color of the agent changes midway through the episode, and (3) a trajectory where the agent is teleported to a random position at the same timestep. The resulting surprise signals for PLDM and DINO-WM are shown in the left panels of Figures~\ref{fig:voe_surprise_pldm} and~\ref{fig:voe_surprise_dinowm}, respectively.

\paragraph{PushT.}
In the \textit{PushT} environment, the agent is controlled by a random policy biased toward interacting with the block. As before, we construct three trajectories: (1) an unperturbed trajectory, (2) a trajectory where the color of the block changes abruptly during the episode, and (3) a trajectory where both the agent and the block are teleported to random positions at the perturbation timestep. The corresponding surprise signals for PLDM and DINO-WM are shown in the center panels of Figures~\ref{fig:voe_surprise_pldm} and~\ref{fig:voe_surprise_dinowm}.

\paragraph{OGBench-Cube.}
In the \textit{OGBench-Cube} environment, the agent follows an expert policy that picks up the cube and places it at a target position. We again consider three trajectories: (1) an unperturbed trajectory, (2) a trajectory where the cube's color changes during the episode, and (3) a trajectory where the cube is teleported to a random position midway through the trajectory. The resulting surprise signals for PLDM and DINO-WM are shown in the right panels of Figures~\ref{fig:voe_surprise_pldm} and~\ref{fig:voe_surprise_dinowm}.

\begin{figure}
    \centering
    \includegraphics[width=\linewidth]{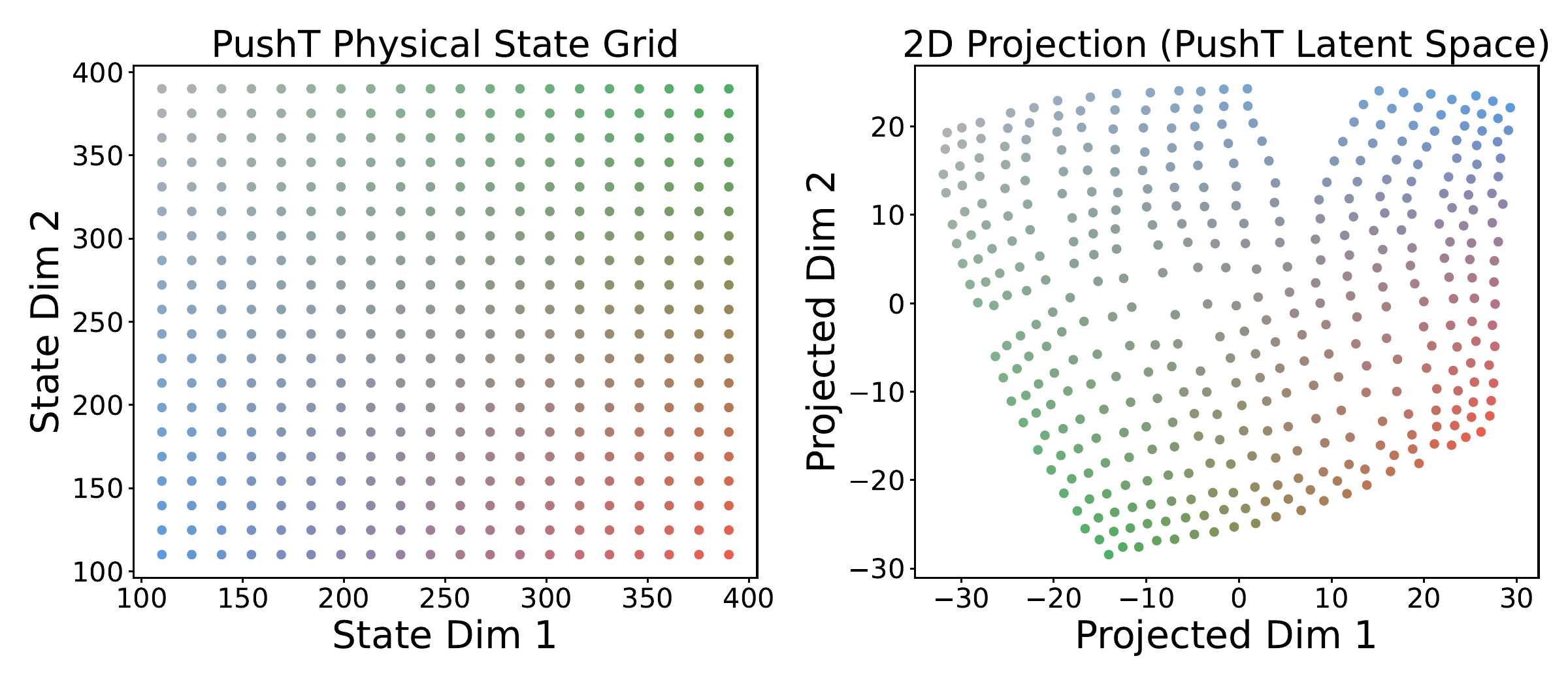}
    \caption{\small \textbf{Visualization of the latent space}  obtained with LeWM for the PushT environment. On the left, the grid of states is obtained by moving the agent and the block in the x-y plane. On the right, the embeddings of these states are visualized using a t-SNE.}
    \label{fig:tsne_pusht_lejepa}
\end{figure}

\section{Ablations.}
\label{appendix:ablations}

\begin{figure}[t]
    \centering
    \includegraphics[width=0.32\linewidth]{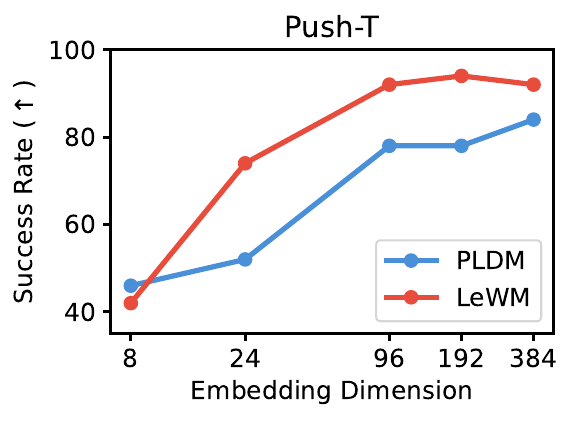}
    \includegraphics[width=0.32\linewidth]{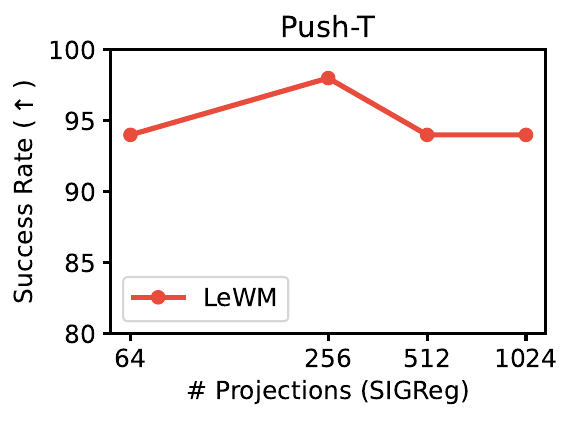}
    \includegraphics[width=0.32\linewidth]{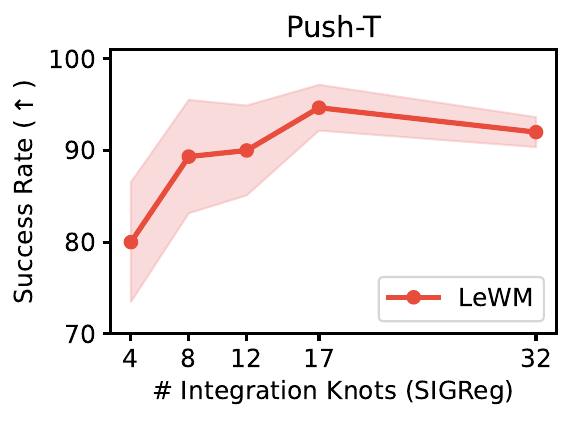}
    \caption{\textbf{Ablation studies of key design choices in LeWM.} \textbf{Left:} effect of the embedding dimension; performance improves with larger embeddings but quickly saturates beyond a certain threshold. \textbf{Center:} effect of the number of random projections used in SIGReg; performance remains stable, indicating that this parameter is not critical. \textbf{Right:} effect of the number of integration knots used to compute the SIGReg loss; results are similarly insensitive to this parameter.}
    \label{fig:ablation-sigreg}
\end{figure}

\paragraph{Training variance.}
To assess the stability of training, we retrain the model using multiple random seeds. As shown in Tab.~\ref{tab:train-var}, the resulting performance exhibits consistently high success rates with low variance across runs, indicating that the training procedure is stable and reproducible.

\begin{table}[h]
\centering
\caption{\textbf{Training Variance.} We report the mean success rate across three training seeds and the corresponding variance, evaluated over the same set of 50 trajectories on Push-T. The goal configuration is reachable within 25 steps, and we allow a planning budget of 50 steps. PLDM exhibits higher variance compared to DINO-WM and LeWM.}
\begin{tabular}{l c}
\toprule
\textbf{Model} & \textbf{Push-T} (SR $\uparrow$)\\
\toprule
DINO-WM & $92.0 \pm 1.63$\\
PLDM & $78.0 \pm 5.0$ \\
LeWM (ours) & $\bm{96.0} \pm \bm{2.83}$\\
\bottomrule
\end{tabular}
\vspace{2mm}
\label{tab:train-var}
\end{table}

\paragraph{Embedding dimensions.}
We study the impact of the embedding dimensionality on performance. As shown in Fig.~\ref{fig:ablation-sigreg}, performance drops when the embedding dimension falls below a certain threshold (around 184), while increasing the dimension beyond this value yields diminishing returns and leads to performance saturation.

\paragraph{Number of projections in SIGReg.}
We study the impact of the number of projections used in SIGReg. As shown in Fig.~\ref{fig:ablation-sigreg}, varying the number of projections has little effect on performance in downstream control tasks. This suggests that the method is largely insensitive to this hyperparameter, and therefore it does not require careful tuning. In practice, this leaves $\lambda$ as the only effective hyperparameter to optimize.

\begin{figure}
    \centering
    \includegraphics[width=0.5\linewidth]{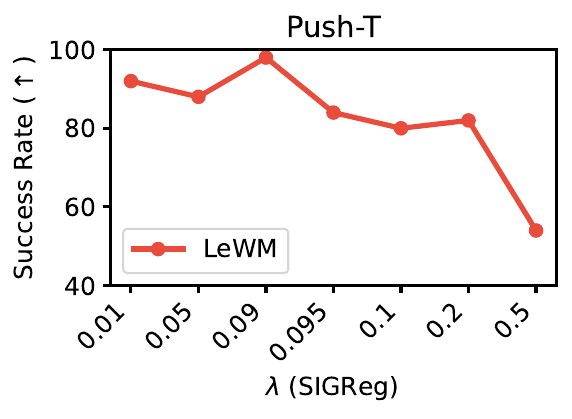}
    \caption{\small \textbf{Effect of the SIGReg regularization weight $\lambda$ on Push-T planning performance.} Success rate remains above 80\% across a wide range of values ($\lambda \in [0.01, 0.2]$), peaking near $\lambda = 0.09$. Performance degrades sharply only at $\lambda = 0.5$, where the regularizer dominates the prediction loss and hinders dynamics modeling. Since $\lambda$ is the only effective hyperparameter of LeWM, the SIGReg loss coefficient is easy to tune via a simple bisection search.}
    \label{fig:abl-lambda}
\end{figure}

\paragraph{Weight of SIGReg regularization.}
We analyze the effect of the SIGReg regularization weight $\lambda$. As shown in Fig.\ref{fig:abl-lambda}, the method achieves high performance across a wide range of values for $\lambda$. In particular, for $\lambda \in [0.01, 0.2]$, the success rate remains above 80\%. This indicates that the approach is robust to the choice of this parameter. Moreover, since $\lambda$ is the only effective hyperparameter, it can be tuned efficiently, for instance via a simple bisection search.

\paragraph{Predictor Size.}
We analyze the effect of the predictor size on performance. As shown in Tab.~\ref{tab:pred-size}, the best results are obtained with a ViT-S predictor. Reducing the predictor to a ViT-T model leads to a drop in performance, while increasing the size to ViT-B does not provide additional gains and slightly degrades performance. This suggests that ViT-S offers the best trade-off between model capacity and optimization stability for this task.

\begin{table}
\centering
\caption{\small Effect of the predictor size on planning performance in the Push-T environment. We report the success rate (SR). The ViT-S predictor achieves the best performance.}
\begin{tabular}{l c}
\toprule
\textbf{pred. size} & \textbf{Push-T} (SR $\uparrow$)\\
\toprule
tiny & $80.67 \pm 6.54$\\
small & $\bm{96.0} \pm \bm{2.83}$\\
base & $86.7 \pm 3.06$\\
\bottomrule
\end{tabular}
\label{tab:pred-size}
\end{table}

\paragraph{Decoder.}
We study the impact of adding a reconstruction loss during training. As shown in Tab.~\ref{tab:decoder-loss}, incorporating a decoder and a reconstruction objective does not improve downstream control performance. In fact, performance slightly decreases compared to the model trained without a decoder. This suggests that the JEPA training objective already captures the information necessary for planning, while the reconstruction loss may encourage the model to encode additional visual details that are not relevant for control.

\begin{table}
\centering
\caption{\small Effect of adding a reconstruction loss during training. We report the success rate (SR) on the Push-T planning task. The model trained without the decoder loss achieves higher performance.}
\begin{tabular}{l c}
\toprule
 & \textbf{Push-T} (SR $\uparrow$)\\
\toprule
LeWM w/o decoder loss & $\bm{96.0} \pm \bm{2.83}$\\
LeWM with decoder loss & $86.0 \pm 7.54$\\
\bottomrule
\end{tabular}
\label{tab:decoder-loss}
\end{table}

\paragraph{Architecture.}
We study the impact of encoder architecture on LeWM performance by replacing the ViT encoder with a ResNet-18 backbone. As shown in Tab.~\ref{tab:encoder-arch}, LeWM achieves competitive performance with both architectures, suggesting that it is agnostic to the choice of vision encoder used during training, though ViT retains a modest advantage.

\begin{table}[h]
\centering
\caption{\small \textbf{Encoder Architecture Effect.} We report the success rate (SR) on the Push-T planning task. LeWM achieves competitive performance across encoder architectures, with ViT holding a slight edge.}
\begin{tabular}{l c}
\toprule
 & \textbf{Push-T} (SR $\uparrow$)\\
\midrule
LeWM ViT & $\bm{96.0} \pm \bm{2.83}$\\
LeWM ResNet-18 & $94.0 \pm 3.27$\\
\bottomrule
\end{tabular}
\label{tab:encoder-arch}
\end{table}

\paragraph{Predictor Dropout.}
We analyze the effect of applying dropout in the predictor during training. As shown in Tab.~\ref{tab:pred-drop}, introducing a small amount of dropout significantly improves downstream control performance. In particular, a dropout rate of $0.1$ achieves the highest success rate, while both lower and higher values lead to worse performance. This suggests that moderate dropout helps regularize the predictor and improves generalization, whereas excessive dropout degrades the quality of the learned dynamics.

\begin{table}
\centering
\caption{\small Effect of predictor dropout during training on Push-T planning performance. We report the success rate (SR). A small amount of dropout ($p=0.1$) yields the best results.}
\begin{tabular}{l c}
\toprule
$p$ & \textbf{Push-T} (SR $\uparrow$)\\
\toprule
$0.0$ & $78 \pm 6.54$\\
$0.1$ & $\bm{96.0} \pm \bm{2.83}$\\
$0.2$ & $85.33 \pm 5.74$\\
$0.5$ & $66.67 \pm 4.11$\\
\bottomrule
\end{tabular}
\label{tab:pred-drop}
\end{table}

\paragraph{Planning Solver.} We compare planning performance across diverse solvers. As shown in Tab.~\ref{tab:solver}.

\begin{table}[h]
\centering
\caption{\small \textbf{Planning Solver Performance.} We report the success rate (SR) on the Push-T planning task.}
\begin{tabular}{l c c}
\toprule
 & \textbf{LeWM} & \textbf{PLDM}\\
\midrule
CEM & $\bm{96.0} \pm \bm{2.83}$ & $78.0 \pm 5.0$\\
SGD ($\beta_1$) & $26 \pm 4.32$  & $4.67 \pm 0.06$\\
RMSProp ($\beta_2$) & $67.33 \pm 2.49$ & $49.33 \pm 8.26$\\
Adam ($\beta_1,\beta_2$) & $84 \pm 7.12$ & $80 \pm 3.27$\\

\bottomrule
\end{tabular}
\label{tab:solver}
\end{table}

\section{Temporal Latent Path Straightening.}
\label{appendix:tmp-straight}

The temporal straightening hypothesis, introduced by \citet{henaff2019perceptual}, posits that we represent complex temporal dynamics as smooth, approximately straight trajectories in our representation spaces. This principle has since found applications beyond neuroscience: \citet{interno2025aigenerated} leverage temporal straightness measured from DINOv2 features to discriminate AI-generated videos from real ones, demonstrating that this geometric property carries a meaningful signal about the nature of the underlying dynamics, and \citet{wang2026temporal} shows it can be beneficial for planning.

\begin{figure}
    \centering
    \includegraphics[width=0.5\linewidth]{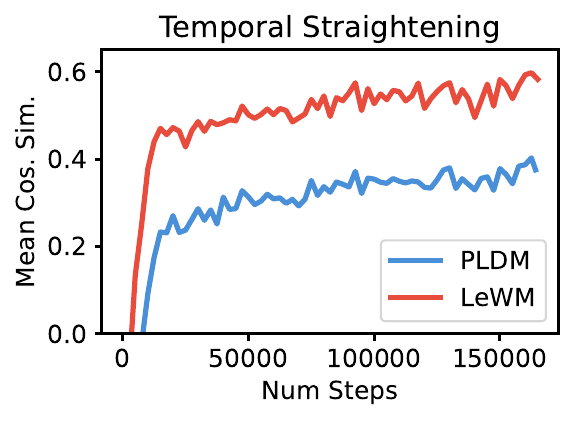}
    \caption{\textbf{Temporal Latent Straightening on Push-T.} Mean cosine similarity between consecutive latent velocity vectors (Eq.~\ref{eq:path-straightening}) over training. Higher values indicate straighter latent trajectories. PLDM explicitly encourages temporal regularity through a dedicated temporal smoothness loss ($\mathcal{L}_{\text{time-sim}}$), yet LeWM achieves substantially straighter latent paths as a purely emergent phenomenon, without any temporal regularization term in its objective.}
    \label{fig:tmp-straight}
\end{figure}

During training on PushT, we record, for curiosity, the temporal straightness of LeWM's latent trajectories. Given a sequence of latent embeddings $\mathbf{z}_{1:T} \in \mathbb{R}^{B \times T \times D}$, we define the temporal velocity vectors as $\mathbf{v}_t = \mathbf{z}_{t+1} - \mathbf{z}_t$. The path straightening measure is defined as the mean pairwise cosine similarity between consecutive velocities:
\begin{equation}
    \mathcal{S}_{\text{straight}} = \frac{1}{B(T-2)} \sum_{i=1}^{B} \sum_{t=1}^{T-2} \frac{\langle \mathbf{v}_t^{(i)},\, \mathbf{v}_{t+1}^{(i)} \rangle}{\|\mathbf{v}_t^{(i)}\| \, \|\mathbf{v}_{t+1}^{(i)}\|}.
    \label{eq:path-straightening}
\end{equation}
A value of $\mathcal{S}_{\text{straight}}$ close to $1$ indicates that consecutive velocities are nearly collinear, meaning the latent trajectory approaches a straight line. Interestingly, we observe that temporal straightening emerges naturally over the course of training without any training term explicitly encouraging it (Fig.~\ref{fig:tmp-straight}). 

We hypothesize that this emerges because SIGReg is applied independently at each time step but not across the temporal dimension, leaving the temporal structure unconstrained. This allows the encoder to converge toward a form of \emph{temporal collapse}, where successive embeddings evolve along increasingly linear paths. Rather than being detrimental, this implicit bias appears to benefit downstream performance, as shown in Fig.~\ref{fig:ctrl-all}. Notably, LeWM achieves higher temporal straightness than PLDM despite having no explicit regularizer encouraging it, whereas PLDM employs a regularizer on consecutive latent states that directly promotes temporal smoothness.

\section{Training Curves}
\label{appendix:train-curves}
\begin{figure}
    \centering
    \includegraphics[width=\linewidth]{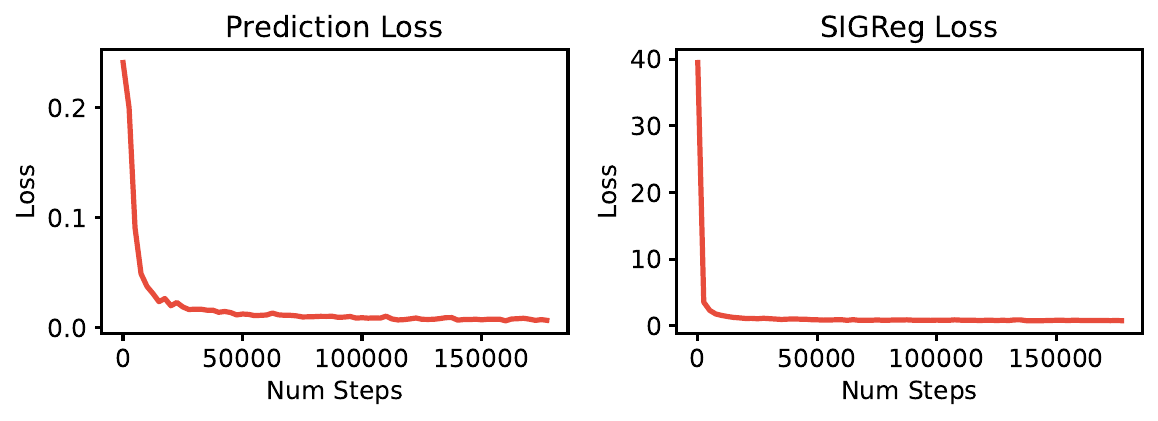}
    \caption{\textbf{Push-T Training curves for LeWM.}}
    \label{fig:lewm-loss}
\end{figure}

\begin{figure}
    \centering
    \includegraphics[width=\linewidth]{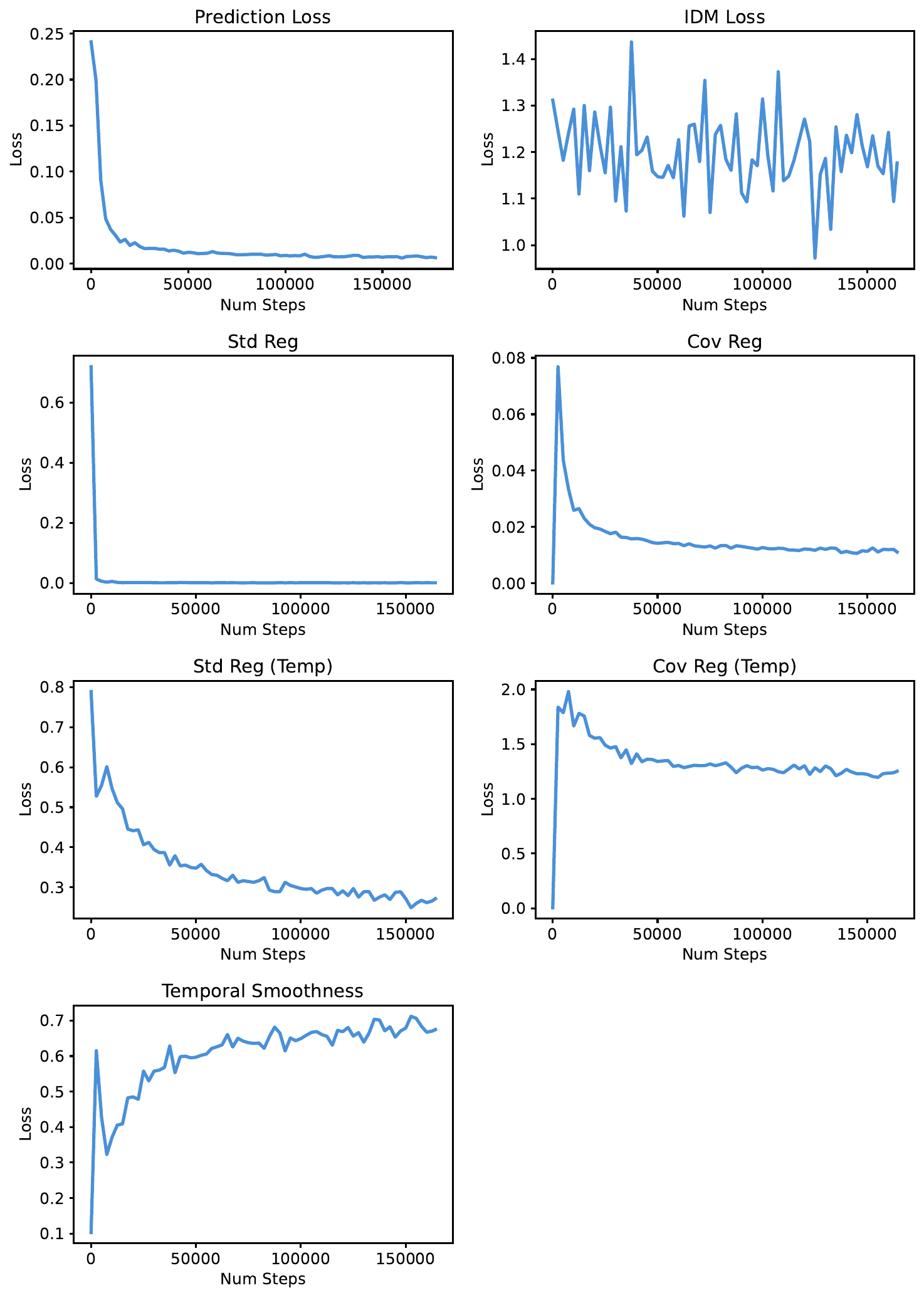}
    \caption{\textbf{Push-T Training curves for PLDM.}}
    \label{fig:pldm-loss}
\end{figure}

We visualize several training curves comparing the optimization dynamics of LeWM (Fig. \ref{fig:lewm-loss}) and PLDM (Fig. \ref{fig:pldm-loss}). In contrast to PLDM, whose objective contains multiple regularization terms, LeWM uses a single regularization term in addition to the prediction loss, making the training dynamics easier to interpret and analyze.

\end{document}